\documentclass[10pt,twocolumn,letterpaper]{article}
\usepackage[pagenumbers]{cvpr}
\usepackage{times,graphicx,amsmath,amssymb,booktabs,multirow,tabularx,microtype,xcolor}
\usepackage{algorithm,algpseudocode}
\definecolor{cvprblue}{rgb}{0.21,0.49,0.74}
\usepackage[pagebackref,breaklinks,colorlinks,allcolors=cvprblue]{hyperref}
\usepackage[capitalize]{cleveref}
\makeatletter
\renewcommand\paragraph{\@startsection{paragraph}{4}{\z@}{6pt plus .5pt minus .5pt}{-1em}{\normalfont\normalsize\bfseries}}
\def\NAT@sort{0}
\def\NAT@cmprs{0}
\makeatother
\def\confName{CVPR}
\def\confYear{2027}

\newtheorem{proposition}{Proposition}
\author{%
  \textbf{Xiyang Zhang\textsuperscript{1,2}},
  \textbf{Yuanhe Tian\textsuperscript{2,3}}\thanks{Corresponding author},
  \textbf{Hongzhi Wang\textsuperscript{1}},
  \textbf{Yan Song\textsuperscript{4}}\\[0.5em]
  {\normalfont
    \textsuperscript{1}Harbin Institute of Technology,\quad
    \textsuperscript{2}Zhongguancun Academy}\\
  {\normalfont
    \textsuperscript{3}Zhongguancun Institute of Artificial Intelligence}\\
  {\normalfont
    \textsuperscript{4}University of Science and Technology of China}\\[0.5em]
  {\normalfont\small
    \texttt{s-zhangxiy24@bza.edu.cn}\quad
    \texttt{tianyuanhe@zgci.ac.cn}}\\
  {\normalfont\small
    \texttt{wangzh@hit.edu.cn}\quad
    \texttt{clksong@gmail.com}}%
}
\title{TRACE: Attribute Execution for Sequential Image Editing}

\makeatletter
\let\TRACE@maketitle\maketitle
\let\TRACE@@maketitle\@maketitle
\let\TRACE@author\@author
\let\TRACE@thanks\thanks
\newcommand{\TRACErestoresupplementtitle}{%
  \let\maketitle\TRACE@maketitle
  \let\@maketitle\TRACE@@maketitle
  \let\@author\TRACE@author
  \let\thanks\TRACE@thanks
  \let\@thanks\@empty
  \def\@title{TRACE: Attribute Execution for Sequential Image Editing\\ Supplementary Material}%
}
\makeatother

\begin{document}\maketitle
\begin{abstract}
Given an image and successive user instructions, sequential image editing should return an updated image that satisfies each new request and preserves earlier requirements that remain active. This lets users develop a result without repeatedly restating or repairing their decisions. The difficulty is that a retrieved image can contain a valid color on an obsolete surface, while the current image may lack the newly requested material. Neither reference selection nor a precise object mask alone specifies how to reconcile attributes occupying the same region. We present TRACE, which separates the continuing attribute from its visual carrier and prioritizes local appearance operations under complete-request checks. Active-goal binding identifies what should persist, carrier selection supplies suitable current or newly generated pixels, and tonal calibration reduces tinting near highlights and shadows. Experiments across frozen editors, independent visual readers, new source images, and complete sequential benchmarks support joint color--material recovery and show how local execution preserves the surrounding scene.
\end{abstract}

\section{Introduction}
Sequential image editing lets a user develop an image through a conversation: given the current image and a new instruction, the system returns an updated image that should also respect earlier decisions. A user might first change a sign to yellow, then ask for a wooden surface, and later revise its shape. This workflow is useful precisely because each request can build on the preceding result. The challenge is deciding what should survive each change. A new color replaces the old color, whereas a material change usually leaves the color requirement intact. Moreover, an image records what the editor produced, which may differ from what the user requested. An earlier edit can fail, and a later edit can undo an earlier success. A sequential editor must therefore track the intended state throughout the conversation and recover valid requirements when they disappear from the image.

Instruction-trained editors improve the fidelity of individual changes~\cite{instructpix2pix,mgie,smartedit,emuedit,step1xedit,qwenimage,fluxkontext}, while multi-turn training and conditioning reduce drift across successive outputs~\cite{mcedit,vincie,mteditflow,editr2,mtopsd,mtedit,chen2026mtforcing}. Memory and reference-selection methods retain entity state and retrieve earlier images that contain useful visual evidence~\cite{canvas,respec,seam}. Yet even correct retrieval leaves an important choice unresolved: which part of that evidence should be reused? In the sign example, an earlier image may show the desired yellow on the old surface. Copying the sign restores yellow but also removes the new wood grain. A precise object mask cannot prevent this regression, because color and material occupy the same pixels. If the earlier color edit failed, the retrieved image may not contain the desired appearance at all. Thus a continuing requirement, its visual evidence, and the surface on which it should appear need separate representations. Color-transfer and color-editing methods already offer ways to change color while retaining structure~\cite{reinhard2001color,coloredit,colorctrl}; sequential recovery additionally requires deciding whether the target is still valid and whether the resulting image satisfies the newer request.

In this paper, we propose TRACE, an attribute executor around a frozen image editor. Its central idea is to reuse an appearance operation whose target and application image are bound separately. \emph{Active-goal binding} follows the instructions to retain complementary requirements, retire revised values, and associate each target with its original request. \emph{Local attribute execution} uses qualified history to specify the desired color, while the current object supplies the spatial samples and surface structure. When the current object lacks the newly requested material, a corrective image supplies a replacement surface for the same local region. Tonal calibration fits the target color center while tapering changes near highlights and shadows. \emph{Complete-request selection} then checks both the earlier and current requests in their original image contexts. A color match alone is insufficient when an earlier instruction also specifies identity or other properties. Successful operations can be retained and applied again after fresh target binding and verification, adapting execution without changing the editor's weights.

We study this principle through color recovery after material changes, where copying historical content can directly conflict with the current request. Matched-proposal experiments separate the effect of the operation from planning, reference retrieval, and additional generation. Tests on new photographs, color revisions, and two independent appearance readers examine transfer and the coexistence of both requested attributes. Complete EdiVal and MagicBrush trajectories further measure instruction following and content consistency under the original benchmark protocols. Together, these experiments show where local attribute execution improves recovery and how it preserves the surrounding scene compared with returning a corrective image in full.

Our contributions are:
\begin{itemize}[topsep=3pt,itemsep=3pt,parsep=0pt]
\item We identify the distinction between a continuing attribute, the historical evidence that specifies it, and the current surface on which it must be realized in sequential image editing.
\item We introduce TRACE, which binds active goals to local appearance operations, acquires a suitable surface when needed, and verifies the combined result against complete requests.
\item We establish the effect of current-sample execution through matched controls, and evaluate joint recovery, tonal calibration, goal revision, and scene preservation across editors and complete sequential benchmarks.
\end{itemize}

\section{Related Work}
\paragraph{Sequential editing and memory.} Multi-turn training and conditioning reduce accumulated drift~\cite{mcedit,vincie,mteditflow,editr2,mtopsd,acedit,mtedit,chen2026mtforcing,anchoredit,freqedit}. Visual and entity memories retain earlier states~\cite{kim2025layerwisememory,canvas,seam}, while ReSpec selects restoration evidence and identity anchors~\cite{respec}. TRACE addresses the subsequent execution problem: applying a valid historical attribute to an object whose other properties have changed.

\paragraph{Appearance and spatial control.} Color transfer and diffusion-based color editing separate appearance guidance from image structure~\cite{reinhard2001color,coloredit,colorctrl}. Attention, inversion, feature, and mask control restrict changes to relevant regions~\cite{sdedit,prompt2prompt,masactrl,pix2pixzero,plugandplay,diffedit,imagic,nulltext,blendedlatent,zone,chen2024mimicbrush,rfinversion,flowedit,kvedit,overpainting,qu2026replan}. These tools control an edit's appearance or location. TRACE additionally resolves which earlier attribute remains valid and how to apply it without restoring an obsolete surface in the same region.

\paragraph{Executable memory and feedback.} Visual programs and retained skills make operations reusable~\cite{gupta2023visual,wang2023voyager}. Recent agentic editors and harness adaptation methods organize frozen-model inference through planning, tools, and feedback~\cite{agentbanana,imagagent,xia2026rrsi,xu2026harness2,yang2026rise,guan2026bluepixel}, supported by question-based visual assessment~\cite{tifa,geneval,viescore,wu2026editreward,huang2026editjudgebias}. TRACE uses a fixed control policy with reusable attribute operations, rebinding their targets and supports and verifying each application as the requested state changes.

\section{Method}
\label{sec:method}

\begin{figure*}[t]
\centering
\includegraphics[width=\textwidth]{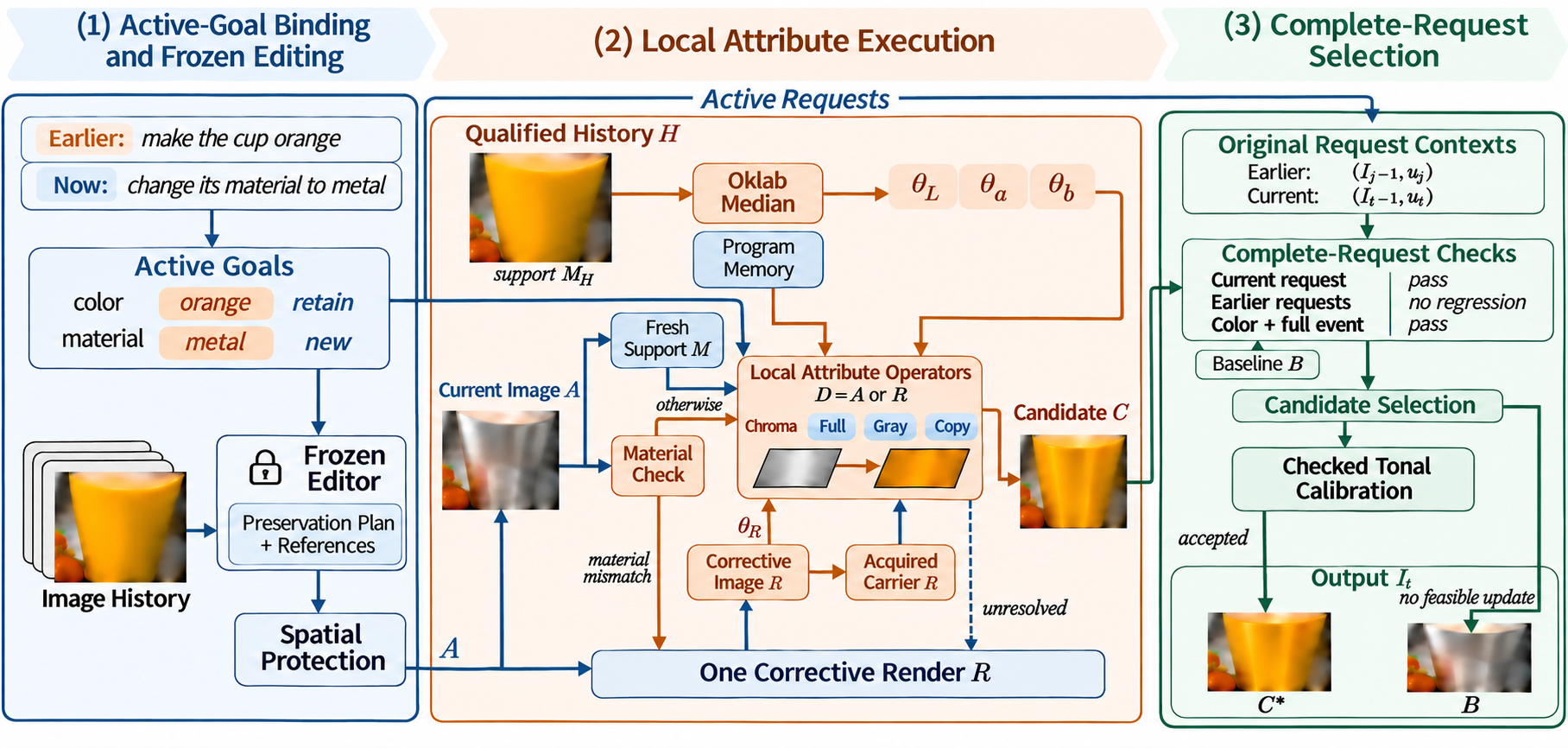}
\caption{TRACE separates the desired attribute from the surface that carries it. Active goals identify what should persist. History supplies color parameters, while current or acquired pixels supply the surface on which they are executed. Complete-request checks select a candidate and verify its tonal calibration. Each attempt uses at most one corrective image for missing color evidence or a missing surface; supported material repair can repeat once when an attribute remains unconfirmed. Local candidates have priority over fully checked whole-image fallbacks. Images in this overview are illustrative.}
\label{fig:method}
\end{figure*}

At turn $t$, TRACE receives instructions $u_{1:t}$ and returned images $I_{0:t-1}$, and produces the next image $I_t$. Figure~\ref{fig:method} shows three components. \emph{Active-goal binding} records the requirements that remain in force and prepares an ordinary editing proposal $A$. \emph{Local attribute execution} recovers a missing color on a suitable object surface, producing a candidate $C$. \emph{Complete-request selection} accepts the candidate only after checking the current and applicable earlier requests. Throughout, the \emph{target} is the appearance to recover, and the \emph{carrier} is the image whose object pixels receive the operation. They may come from different turns.

\subsection{Active-goal binding}
\label{sec:goal-binding}
Active-goal binding separates the user's requested state from the editor's rendering. A goal $g=(e,a,v,\tau,s)$ records the entity, attribute, target value, introducing turn, and original clause. A text-only interpreter updates the requested state $S_t$ from $S_{t-1}$ and the new instruction $u_t$; $\mathcal G_t=\mathsf{Live}(S_t)$ contains the active goals. Revising an attribute retires its old value. Changing a different attribute retains the earlier goal when the entity and its scope are unchanged. A failed rendering leaves a goal unmet, so it remains available for later recovery.

Each atomic goal is linked to the complete request that introduced it. For example, replacing an object with a yellow sign specifies both an identity and a color; recovering yellow alone is insufficient. An earlier request is checked while all its requirements remain active. The implementation supports the current request and up to two complete earlier requests, and checks replacement, membership, and reference ambiguity before carrying a target forward. Unsupported cases retain the ordinary edit.

We use ReSpec's preservation plan and ordered image references~\cite{respec} to produce the ordinary proposal $Y_t$, adding the continuing goals while giving the current instruction precedence. For local requests, separately grounded SAM2.1 masks~\cite{sam2} can restrict unrelated changes when input and proposal supports agree. This optional spatial restriction uses the complete-request check in Sec.~\ref{sec:request-selection}; its result is $A$. The supplement gives the support and alignment rules.

\subsection{Carrier-qualified attribute execution}
\label{sec:local-execution}
Local execution separates two choices: the color parameters $\theta$ and the carrier image $D$ on which they act. Given an object support $M$, all parametric candidates share the form
\begin{equation}
C_{\theta,D}(p)=\begin{cases}\mathsf{Encode}\!\left(T_\theta(\phi(D(p)))\right),&p\in M,\\A(p),&p\notin M.\end{cases}
\label{eq:unified-execution}
\end{equation}
Here $\phi$ maps RGB to Oklab, $T_\theta$ is fitted to the carrier samples on $M$, and encoding includes gamut mapping. Regional copying uses the identity operation. Qualified history can supply $\theta$ while the current image supplies $D=A$. If color evidence is missing, a corrective image supplies $\theta$; if the current material is missing, it instead supplies the carrier $D$. Thus the same local operator addresses missing appearance evidence and an unsuitable surface. Active goals determine which target is valid, and complete-request checks determine whether its realization can be returned.

\paragraph{Choosing the carrier.} For a supported atomic material request with one continuing color goal on the same object, an observer describes the visible material from the image and a neutral category such as ``cup.'' The requested value is withheld during observation and compared with the description afterward; we refer to this as a \emph{target-blind} reading. A known mismatch invokes carrier acquisition; matching, uncertain, or unsupported readings retain ordinary color recovery. This separates a missing attribute from a missing surface before choosing an operation.

\paragraph{Executing a continuing attribute.} In the ordinary route, a target-blind observer describes the object's visible body color. A known mismatch identifies at most one target in request order and its unique current support $M$. Historical evidence must have the same entity, the requested color, and its own qualified support $M_H$. Separate supports bind the target and carrier to their respective images. The latest eligible frame $H$ supplies coordinatewise color medians. Using the Oklab transform~\cite{ottosson2020oklab},
\begin{equation}
\theta=\operatorname{med}_{p\in M_H}\phi(H(p)),\quad m=\operatorname{med}_{p\in M}\phi(A(p)).
\label{eq:program-estimation}
\end{equation}
For $x_p=(L_p,a_p,b_p)=\phi(A(p))$, the chroma family~\cite{reinhard2001color} applies
\begin{equation}
T^\chi_\theta(x_p)=\bigl(L_p,a_p+\theta_a-m_a,b_p+\theta_b-m_b\bigr).
\label{eq:chroma-program}
\end{equation}
The historical surface is absent from this expression. Current pixels retain their positions and continuous lightness; history supplies only the color displacement. Fixed-lightness radial gamut mapping and RGB encoding produce the candidate, with exact copies of $A$ outside $M$.

The full family also changes lightness:
\begin{equation}
\begin{aligned}
L'_p&=\theta_L+s(L_p-m_L),\\
(a'_p,b'_p)&=(a_p,b_p)+(\theta_a,\theta_b)-(m_a,m_b).
\end{aligned}
\label{eq:program-full}
\end{equation}
The largest $s\in[0,1]$ keeping lightness in $[0,1]$ retains as much current lightness contrast as this family permits. Chromatic targets try chroma before the full family; black and white start with the full family, and gray first tries zero chroma at current lightness. Family choice therefore reflects the requested attribute rather than applying one historical replacement to every target.

Accepted color operations are stored with their goal and source bindings. Reuse re-establishes the support and repeats the complete-request checks on the new image. If historical or retained parameters do not produce an accepted update, one corrective render $R$ supplies new color parameters sampled within the current support. TRACE applies these parameters to the current carrier $D=A$ and also considers copying the corresponding region from $R$. Both candidates retain $A$ outside $M$. If neither passes, selection proceeds to the checked fallback below. Thus a corrective image can guide color recovery without replacing the current surface.

\paragraph{Acquiring a missing carrier.} When the current material is incorrect, the corrective image instead supplies a new surface. First, color recovery without corrective generation supplies fallback $P$, which equals $A$ if no operation passes. TRACE then compiles the live goals into a corrective instruction: the complete current request takes precedence, while the neutral category receives the continuing color and required material texture and reflectance. Earlier referring expressions are excluded from this rendering instruction; their complete original requests remain in verification. We bind support $M$ on $A$ and use the same fixed window on $R$. The draw $R$ supplies two candidate types: copying its region into $A$, or executing eligible historical color on the acquired carrier $D=R$. Both copy $A$ exactly outside $M$. If neither type passes, the local route retains $P$. Color recovery uses at most one corrective image; supported material acquisition can use a second attempt as described below.

\paragraph{Calibrating color without tinting tonal endpoints.} A uniform chroma displacement can tint the carrier's highlights and shadows. After selecting a parametric operation, TRACE tests a tapered refinement when one continuing color request and the current atomic material request refer to the same object. On the selected carrier $D$, let $\mu=\operatorname{med}_{p\in M}L_p$, where $(L_p,a_p,b_p)=\phi(D(p))$. Fits require $0<\mu<1$ and $0<\theta_L<1$. Using the original carrier lightness throughout, define
\begin{equation}
w_\mu(L)=\begin{cases}L/\mu,&L\leq \mu,\\(1-L)/(1-\mu),&L>\mu.\end{cases}
\label{eq:taper-weight}
\end{equation}
The refined chroma is $(a'_p,b'_p)=(a_p,b_p)+w_\mu(L_p)d$. Its displacement vanishes at both tonal endpoints and is largest at the carrier median. Simply setting $d=\theta^{ab}-\operatorname{med}(a,b)$ need not attain the target median because the weights differ across samples. Instead, for each chroma coordinate $c\in\{a,b\}$, we solve
\begin{equation}
d_c=\underset{d:\ \operatorname{med}_{p\in M}(c_p+w_\mu(L_p)d)=\theta_c}{\arg\min}|d|.
\label{eq:calibrated-chroma}
\end{equation}
Chroma operations retain $L_p$. When an interior anchor $0<\tau<1$ exists, full operations use the increasing piecewise-linear map through $(0,0)$, $(\mu,\tau)$ and $(1,1)$, with $\tau$ fitted to the target lightness median. The supplement derives the unique minimum-norm chroma fit and the even-sample lightness correction. Degenerate or infeasible fits retain the selected operation. The refined image uses the same support, target, gamut mapping and exterior pixels. After the initial candidate selection, the calibrated image replaces its parent only if fresh color and complete-request checks pass. Calibration adds no editor draw.

\subsection{Complete-request selection}
\label{sec:request-selection}
The final component checks whether color recovery and the newer edit coexist. It evaluates each candidate against the full current request and the applicable earlier requests, using the image that originally preceded each request to resolve its meaning. A rejected candidate advances the search. These checks test whether a correct-looking attribute coexists with the intended object and the other requested properties.

An observer assesses complete requests as $T$, $F$, or $U$, represented by $J(T)=[1,1]$, $J(F)=[0,0]$, and $J(U)=[0,1]$. The intervals represent possible binary truth values: an unknown answer can be either true or false. Let $c$ denote the current request and $\mathcal H_t$ the applicable earlier requests. The comparison image is $B=A$ for color recovery and $B=P$ for carrier acquisition. With interval endpoints $\ell,u$, a candidate $C$ must satisfy
\begin{equation}
c(C)=T,\qquad \ell\bigl(h_j(C)\bigr)\geq u\bigl(h_j(B)\bigr),\quad j\in\mathcal H_t.
\label{eq:goal-dominance}
\end{equation}
Thus an earlier positive request must remain positive, and an uncertain request must become positive before the candidate is accepted. The repaired color and its complete introducing request must both pass. Checking the original request tests identity and other conditions in addition to the color target.

Search first tries retained operations and then qualified history within each color family, stopping at the first accepted update. If a corrective image is needed, TRACE ranks accepted candidates by request satisfaction and preservation of the carrier's lightness pattern. Carrier acquisition also uses the neutral material reading to prefer a suitable new surface. All acquired candidates must pass the same complete-request checks; if none passes, the route retains its previous result. Exact ranking objectives and tie rules appear in the supplement.

Local candidates have priority. If local recovery leaves the ordinary proposal unchanged ($C=B=A$) and a corrective image $R$ is already available, TRACE checks $R$ itself for the target color and every applicable complete request. Only an all-positive result permits full-image replacement; otherwise $A$ remains. This fallback expands the editing scope only after local execution makes no progress, using the same single draw.

For wood or metal requests that triggered carrier acquisition, TRACE checks both attributes on the first returned result $C_1$. If either reading is $F$ or $U$, a second draw from the same $A$ and seed requests the continuing color as a wood stain or anodized finish, retaining visible grain or reflectance. It uses the same target, support, local execution, calibration, and fallback. The result $C_2$ replaces $C_1$ only when both neutral readings and every applicable complete request are $T$; otherwise $C_1$ remains. The corrective budget is at most two on this route and one elsewhere.

\begin{proposition}[Locality and checked progress]
Every local candidate preserves encoded pixels outside $M$. With correct bindings and sound judgments, admission preserves checked requests satisfied on $A$; repairing a known color or current-material failure increases their total satisfaction by at least one.
\end{proposition}
Before gamut mapping and encoding, affine chroma transfer preserves pairwise sample differences; its full family matches the target median with maximal feasible lightness contrast. Tapered refinement attains the chroma medians with minimum displacement in its fixed family while preserving tonal endpoints and lightness order. The supplement gives the proofs, numerical treatment, and observation schedule.

\begin{table}[t]
\centering\small
\setlength{\tabcolsep}{4pt}
\begin{tabular}{lrrr}
\toprule
 & Native & \multicolumn{2}{c}{Target-blind Pair /112} \\
\cmidrule(lr){3-4}
Method & Both /108 & 8B & GLM \\
\midrule
Raw editor & 62 & 13 & 9 \\
Full prompt & 78 & 19 & 10 \\
All history & 83 & 28 & 22 \\
ReSpec & 79 & 17 & 8 \\
Adapted SEAM & 62 & 13 & 9 \\
BM25 history & 78 & 19 & 10 \\
CLIP retrieval & 82 & 25 & 16 \\
RISE-Agent (adapted) & 66 & 16 & 17 \\
TRACE & \textbf{96} & \textbf{41} & \textbf{45} \\
\bottomrule
\end{tabular}
\caption{Joint recovery on all 56 color--material conversations at two seeds. Native Both checks complete current and applicable earlier requests (108 pairs); blind Pair checks both visible attributes (112 outputs). Methods share first outputs and generate separate material proposals, with spatial restriction disabled. All TRACE components are enabled.}
\label{tab:prior-methods}
\end{table}

\section{Experiments}
\subsection{Experimental setup}
\paragraph{Benchmarks and editors.} We evaluate all 572 three-turn EdiVal sequences (1,716 edits per controller)~\cite{edival}, and all 1,053 independent MagicBrush edits and 535 multi-turn sessions~\cite{magicbrush}. The primary editor is frozen FLUX.2 [klein] 4B~\cite{flux2klein}, with $512^2$ outputs and its four-step sampler. Qwen-Image-Edit-2511~\cite{qwenimage} provides a second backbone, using $1024^2$ outputs and 40 steps. Qwen3-VL-32B~\cite{qwen3vlreport} supplies goal interpretation, planning, and controller checks. We keep editor weights fixed and match each backbone's sampling settings across methods. The supplement lists guidance, seeds, model roles, and execution details.

\paragraph{Comparisons.} Raw uses the preceding image and current request; Full prompt also concatenates all instructions. All history supplies those instructions with the current image first and earlier images chronologically. ReSpec retains preservation planning and ordered references~\cite{respec}. Adapted SEAM additionally tests entity-state memory and retrieval on the color--material task and EdiVal~\cite{seam}. BM25 and CLIP select historical references by lexical and frozen image--text relevance, respectively~\cite{robertson2009bm25,clip}. RISE-Agent~\cite{yang2026rise} uses the same 32B reasoner and 4B editor with its released current-request-only turnwise policy, toolset and refinement budgets. It receives the common first output and current material request; earlier requests enter evaluation only. Matched internal controls isolate execution; the supplement also compares released ColorCtrl~\cite{colorctrl}.

\paragraph{Metrics.} EdiVal retains released GroundingDINO/Qwen2-VL-7B instruction checks~\cite{liu2023groundingdino,wang2024qwen2vl}, DINOv3 consistency~\cite{simeoni2025dinov3}, and HPSv3 reward~\cite{hpsv3}. Cumulative $\mathrm{IF}_t=N^{-1}\sum_i\prod_{k\leq t}s_{i,k}$ records issue-time success; $O=\sqrt{\overline{\mathrm{IF}}\,\overline{\mathrm{CC}}}$. Here $s_{i,k}$ is the native instruction-success bit, and bars average the three turns. The same 38 positions affected by content-policy refusals remain in all denominators. MagicBrush retains L1, L2, CLIP-I, DINO and CLIP-T~\cite{clip,caron2021dino}.

\paragraph{Continuing requirements.} We also evaluate complete active requests on later outputs, using the image and instruction that introduced each request. This measures recovery that cumulative IF cannot credit after an earlier failure. Paired 95\% bootstrap intervals use 10,000 draws grouped by source image, keeping repeated seeds and turns together. The supplement reports eligibility and the source-disjoint evaluation.

\paragraph{New sources and goal revisions.} COCO120 contains 120 annotation-selected photographs~\cite{lin2014coco}, balanced across eight target categories and eight colors, with successive color and material requests. A separate 32-source study adds a third-turn color revision. Sixteen eight-turn conversations compare active goals, accumulated goals without retirement, and recent-instruction-only editing on common valid targets. Policies follow their own returned images; shared-proposal controls are identified separately. The supplement gives source selection and the complete requests.

\begin{table*}[t]
\centering\small
\setlength{\tabcolsep}{7pt}
\begin{tabular*}{\textwidth}{@{\extracolsep{\fill}}lrrrrrrrr}
\toprule
& \multicolumn{3}{c}{Turn 2} & \multicolumn{4}{c}{Turn 3} & Overall \\
\cmidrule(lr){2-4}\cmidrule(lr){5-8}\cmidrule(lr){9-9}
Controller & IF$_2$ & mIF$_2$ & CC$_2$ & IF$_3$ & mIF$_3$ & CC$_3$ & VQ$_3$ & $O$ \\
\midrule
Raw editor & 42.83 & 65.73 & 92.28 & 26.92 & 65.73 & 89.49 & 6.49 & 64.93 \\
Full prompt & 41.26 & 63.64 & 92.16 & 22.90 & 58.74 & 89.46 & 6.40 & 63.57 \\
All history & 39.69 & 61.19 & \textbf{93.55} & 22.38 & 57.52 & \textbf{92.19} & 6.13 & 63.52 \\
ReSpec & 47.38 & 71.68 & 91.83 & 29.72 & 68.01 & 88.64 & 6.56 & 66.49 \\
Adapted SEAM & 41.96 & 64.51 & 92.03 & 27.27 & 66.78 & 89.15 & 6.47 & 64.78 \\
BM25 history & 41.26 & 63.64 & 92.16 & 23.95 & 60.49 & 89.57 & 6.50 & 63.84 \\
CLIP retrieval & 40.73 & 62.76 & 92.79 & 22.90 & 59.62 & 90.88 & 6.27 & 63.68 \\
Core (internal) & \textbf{48.25} & \textbf{72.90} & 91.60 & 30.77 & 68.71 & 88.37 & \textbf{6.60} & 66.87 \\
TRACE & 48.08 & 72.73 & 91.64 & \textbf{31.12} & \textbf{69.41} & 88.44 & \textbf{6.60} & \textbf{66.93} \\
\bottomrule
\end{tabular*}
\caption{Complete EdiVal trajectories with Klein: all 572 sequences and 1,716 edits per controller, each following its own outputs. IF is cumulative; mIF checks the current instruction; CC measures content consistency; VQ is HPSv3 reward. All columns are higher-is-better; IF, mIF, CC, and $O$ are percentages. Core includes planning and active goals without spatial or local execution. Full per-turn and cross-editor results appear in the supplement.}
\label{tab:edival-main}
\end{table*}

\subsection{Comparison on continuing color and new material}
From EdiVal, we take all first-turn requests specifying a color outcome: 54 color edits and two replacements, including the five first-turn refusals. A second request changes the same object's material to metal, or wood when the source already specifies metal. The primary comparison uses all 56 conversations at Klein seeds 42/43. Methods share each actual first-turn output but generate separate material proposals with the original sampler; spatial restriction is disabled for this controlled task. Operator controls fix the material proposal $A$; ColorCtrl acts on the same $A$ at seed 42.

Native checks use Qwen2-VL-7B to evaluate each complete applicable request. The two replacement requests also specify a material superseded at turn two, leaving 54 applicable earlier requests per seed; Current and appearance retain all 56 outputs. Separate Qwen3-VL-8B and GLM-4.1V-9B readers~\cite{glm2025thinking} describe visible body color and material without targets, history, method identity, or controller decisions. Targets are bound afterward. Native Both assesses complete contextual requests; blind Pair requires both categorical appearance labels, with unknowns counted as unsuccessful. The supplement checks intended-attribute agreement on human-annotated MagicBrush requests and target images, independently of TRACE outputs.

\paragraph{Joint recovery.} TRACE gives the strongest joint recovery among the external controllers under complete requests and both independent readers (Table~\ref{tab:prior-methods}). History conditioning helps retention but can also supply an older surface. Executing qualified color on current samples improves its coexistence with the new material; the accompanying data include individual-attribute counts and paired intervals.

\subsection{Complete sequential benchmarks}

TRACE achieves the highest final instruction following and combined EdiVal score among the compared controllers (Table~\ref{tab:edival-main}). All history favors content consistency, while Core already provides strong planning and active-goal conditioning. Matched-proposal comparisons below isolate local attribute execution.

\begin{table}[t]
\centering\small
\setlength{\tabcolsep}{2.2pt}
\begin{tabular}{lrrrrr}
\toprule
Controller & L1 $\downarrow$ & L2 $\downarrow$ & CLIP-I $\uparrow$ & DINO $\uparrow$ & CLIP-T $\uparrow$\\
\midrule
\multicolumn{6}{l}{\textit{All 1,053 independent edits}}\\
Context baselines & .1116 & .0359 & .9099 & .8522 & \textbf{.2910}\\
TRACE & \textbf{.1111} & \textbf{.0357} & \textbf{.9101} & \textbf{.8525} & \textbf{.2910}\\
\midrule
\multicolumn{6}{l}{\textit{All 535 own-history final images}}\\
Raw editor & .1362 & .0486 & .8777 & .7901 & \textbf{.2925}\\
Full prompt & .1347 & .0474 & .8825 & .7996 & .2910\\
All history & \textbf{.1251} & \textbf{.0434} & \textbf{.8860} & \textbf{.8065} & .2885\\
ReSpec & .1313 & .0453 & .8842 & .8032 & .2921\\
BM25 history & .1346 & .0474 & .8827 & .7987 & .2913\\
CLIP retrieval & .1292 & .0449 & .8851 & .8043 & .2901\\
TRACE & .1313 & .0455 & .8831 & .8015 & .2920\\
\bottomrule
\end{tabular}
\caption{Complete MagicBrush metrics with Klein. Independent edits reset to supplied inputs, giving the six context baselines identical outputs. Own-history sessions consume each controller's returned images. L1/L2 measure pixel error, CLIP-I/DINO image similarity, and CLIP-T text alignment; bold marks the best values and ties within each panel.}
\label{tab:magicbrush-main}
\end{table}

MagicBrush measures agreement with designated reference images alongside alignment with the final instruction (Table~\ref{tab:magicbrush-main}). On independent edits, checked spatial protection improves pixel and image-similarity scores. On own histories, All history gives the strongest image-reference agreement, while TRACE lowers pixel errors relative to Raw and Full prompt and improves final-text similarity over All history. These measures expose complementary preservation and instruction-following tradeoffs.

\begin{figure}[t]
\centering
\includegraphics[width=\columnwidth]{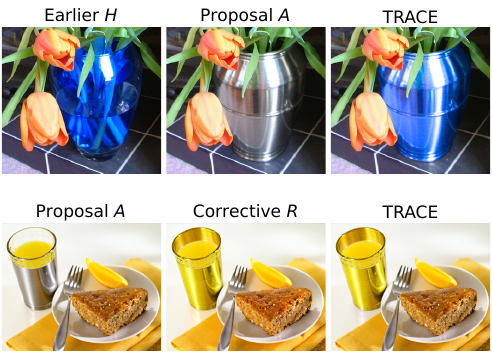}
\caption{Two attribute-recovery routes with identical crops within each row. Top: earlier blue glass $H$, current metal $A$, and TRACE. Bottom: partly glass $A$, acquired metal $R$, and TRACE's local composition.}
\label{fig:material-case}
\end{figure}

\subsection{Analysis of attribute execution}
\paragraph{Matched comparisons.} Core combines ReSpec planning with active-goal prompting, without spatial restriction or local execution. In the controlled experiments, Core is the common material proposal $A$; each operator starts from this same image, history, and target. This design isolates execution from changes in the proposal or subsequent trajectory. TRACE improves joint appearance over Core under both readers on both cohorts; all four paired intervals are positive. Full counts and intervals appear in the supplement.

\paragraph{Spatial support.} Shared-proposal controls show that complete-request checks preserve joint successes and improve consistency, whereas current-only checks can admit historical regressions (supplement).
\paragraph{Attribute execution versus pixel restoration.} Applying historical chroma to current samples outperforms historical pixel restoration in joint request satisfaction and preservation of the newer material (Table~\ref{tab:execution-ablations}). The operations use the same eligible history and fresh supports, isolating the carrier of the update: copying restores an old surface, whereas attribute execution retains the current one. The vase in Figure~\ref{fig:material-case} illustrates why these carriers differ: its historical blue is valid, but its transparent surface is obsolete.

\begin{table}[t]
\centering\small
\setlength{\tabcolsep}{3.0pt}
\begin{tabular}{lrrrr}
\toprule
& Native & Native & \multicolumn{2}{c}{Blind Pair}\\
Method & Both & Current & 8B & GLM\\
$n$ & 108 & 112 & 112 & 112\\
\midrule
Core & 87 & \textbf{110} & 22 & 18\\
Historical pixels & 79 & 92 & 21 & 18\\
Historical chroma & 93 & 109 & 26 & 29\\
\midrule
\multicolumn{5}{l}{\textit{Affine color-recovery controls}}\\
No corrective image & 93 & \textbf{110} & 28 & 28\\
Current-only admission & 92 & \textbf{110} & 30 & 31\\
\midrule
\multicolumn{5}{l}{\textit{Carrier acquisition $\times$ tonal calibration}}\\
Color executor (affine) & 94 & \textbf{110} & 34 & 35\\
Color executor (calibrated) & 93 & 109 & 37 & 37\\
Carrier executor (affine) & \textbf{97} & \textbf{110} & 37 & 42\\
Carrier executor (calibrated) & 96 & 109 & \textbf{41} & 44\\
\midrule
Acquired copy (affine) & 95 & \textbf{110} & 38 & 40\\
Uncalibrated taper & 96 & 109 & 40 & 41\\
Color + half-step & 94 & \textbf{110} & 38 & 34\\
Checked whole corrective & 94 & 109 & 40 & 43\\
Whole corrective (calib.) & 96 & 108 & 40 & \textbf{45}\\
TRACE & 96 & 109 & \textbf{41} & \textbf{45}\\
\bottomrule
\end{tabular}
\caption{Matched execution on 112 common proposals. The central factorial varies acquisition and calibration on the same $A$, with at most one corrective image. TRACE additionally retries unconfirmed attributes with a surface-finish description. Acquired copy omits color execution on acquired pixels; uncalibrated taper omits median fitting; half-step halves the affine color displacement. Whole and Checked whole return the first corrective draws globally, the latter after color and complete-request checks. Unknowns remain in the denominators; full counts appear in the supplement.}
\label{tab:execution-ablations}
\end{table}

\paragraph{Corrective evidence and admission.} Within the affine color executor, removing corrective evidence lowers blind Pair while leaving Native Current unchanged. Replacing complete-request checks with current-only admission lowers both Native Both and blind Pair (Table~\ref{tab:execution-ablations}).

\begin{figure}[t]
\centering
\includegraphics[width=\columnwidth]{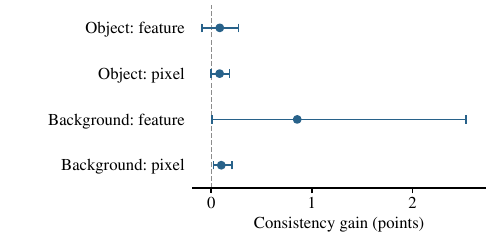}
\caption{Context preservation of the single-proposal executor against Checked whole corrective on all 120 COCO sources. Points show paired executor-minus-control differences with 95\% source-bootstrap intervals, using shared first outputs as references. Objects cover 103 sources; background feature and pixel scores cover 120 and 119. All source clusters enter resampling. The control applies the same request checks before whole-image replacement.}
\label{fig:local-results}
\end{figure}

\paragraph{Using a corrective image locally.} Under the same draw and complete-request checks, the single-proposal executor recovers more joint requirements on the controlled cohort; checked whole-image replacement recovers slightly more COCO pairs. Local execution better preserves the background (Figure~\ref{fig:local-results}). Adaptive refinement revisits outputs whose color or material remains unconfirmed.

\paragraph{Carrier acquisition and calibration.} The four central rows of Table~\ref{tab:execution-ablations} vary carrier acquisition and tonal calibration within the local executor. Acquisition addresses missing material, while calibration controls how color is applied to the selected surface. Their combination gives the strongest joint appearance within this single-proposal factorial. Comparing acquired copying with attribute execution further tests whether the new surface also needs its historical color restored. The final adaptive refinement improves joint recovery under the unchanged attribute and complete-request checks. The supplement reports computational costs and end-to-end timing.

\begin{figure}[t]
\centering
\includegraphics[width=\columnwidth]{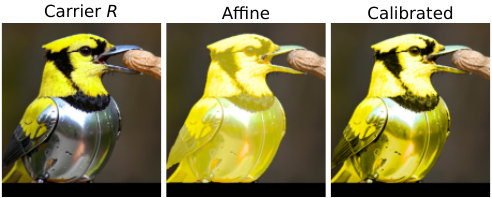}
\caption{Tonal calibration with the same carrier, target and support after requests for yellow, then metal. Affine execution lightens dark head and neck markings; calibration retains them while recovering body color. All panels use the same crop.}
\label{fig:calibration-case-main}
\end{figure}

\paragraph{Operation strength.} Calibrated and uncalibrated taper use identical selected operations, carriers and editor draws, isolating median fitting. Median fitting adds blind Pair successes under both appearance readers (Table~\ref{tab:execution-ablations}). A fixed half-step shifts the readers in opposite directions; simply weakening execution differs from fitting the target median. The supplement reports additional strengths. Figure~\ref{fig:calibration-case-main} shows how endpoint preservation retains dark markings during body-color recovery.

\paragraph{Transfer across samplers.} The local executor improves joint appearance over Core under both readers with another Klein seed and with Qwen. These controls isolate local execution. Stronger Qwen proposals leave fewer color failures to repair. The supplement reports complete counts and Qwen trajectories.

\subsection{Transfer and goal-state analysis}
\paragraph{New-source transfer.} COCO120 tests the fixed executor on new photographs (Table~\ref{tab:new-sources-lifetime}). TRACE improves joint appearance over the external controllers under both readers. Calibrated color execution handles suitable surfaces; acquisition and adaptive refinement address the remaining attribute failures. The source-disjoint three-turn cohort then revises color after material editing, with each method following its own outputs. TRACE retains more current requests than unconditional whole rendering, which recovers more joint requirements. The supplement reports per-turn and context measurements.

\paragraph{Goal lifetime.} Sixteen eight-turn conversations compare active goals with accumulated goals that retain superseded values, using the same planner and affine color executor. Recent-instruction-only editing provides a further reference. Active goals improve complete-request satisfaction over retaining superseded values. The two appearance readers also favor active goals, although their smaller differences indicate that correct goal tracking alone does not ensure visual realization. Policies follow their own outputs under the same valid requests. The supplement reports complete counts, paired intervals, and local/global controls.

\begin{table}[t]
\centering\small
\setlength{\tabcolsep}{3.4pt}
\begin{tabular}{lrrrr}
\toprule
&\multicolumn{2}{c}{Native}&\multicolumn{2}{c}{Blind Pair}\\
\cmidrule(lr){2-3}\cmidrule(lr){4-5}
Policy & Current & Both & 8B & GLM\\
\midrule
\multicolumn{5}{l}{\textit{New COCO sources: 120 two-turn conversations}}\\
Raw editor & 111 & 52 & 3 & 3\\
Full prompt & 109 & 73 & 16 & 14\\
All history & 102 & 68 & 17 & 12\\
ReSpec & 112 & 63 & 13 & 10\\
BM25 history & 109 & 73 & 16 & 14\\
CLIP retrieval & 107 & 73 & 16 & 13\\
RISE-Agent (adapted) & \textbf{115} & 62 & 8 & 4\\
Core (internal) & 113 & 67 & 8 & 8\\
Color executor (calibrated) & 112 & 77 & 21 & 23\\
Local-only execution & 112 & 76 & 22 & 24\\
Single proposal & 112 & 78 & 24 & 26\\
TRACE & 113 & \textbf{79} & 25 & \textbf{27}\\
Checked whole corrective & 112 & 78 & 25 & 26\\
Whole corrective render & 93 & \textbf{79} & \textbf{27} & 25\\
\midrule
\multicolumn{5}{l}{\textit{Independent sources, Klein: both later turns, $n=64$}}\\
Full prompt & 58 & 36 & 16 & 16\\
ReSpec & \textbf{61} & 37 & 17 & 21\\
Color executor (calibrated) & \textbf{61} & 36 & 16 & 18\\
Whole corrective render & 56 & \textbf{46} & \textbf{22} & \textbf{22}\\
TRACE & \textbf{61} & 37 & 16 & 18\\
\bottomrule
\end{tabular}
\caption{New-source evaluation: all 120 COCO outputs and all 64 later outputs from 32 independent three-turn conversations. Local-only permits regional updates; Single proposal also enables the checked whole-image fallback; TRACE adds the conditional second material proposal. These controls distinguish local execution from a larger search budget. Three-turn policies share their first output and then follow their own histories. Bold marks column maxima and ties. Full attribute, per-turn, and context results appear in the supplement.}
\label{tab:new-sources-lifetime}
\end{table}

\subsection{Case study}

The two cases in Figure~\ref{fig:material-case} require different routes. The vase's requested metal surface already exists, but blue is missing. History supplies blue parameters; using its pixels would also restore glass and visible stems. Execution instead acts on the current opaque surface and reflection bands. For the yellow cup, color recovery alone cannot replace the glass upper body. A corrective draw supplies a metallic body and rim, while local composition retains the surrounding meal from the current proposal. Carrier qualification selects the appropriate route before checking both attributes together. The contrast shows why a mask alone cannot determine the update: the same object region may need either color transfer or a newly generated surface.

\section{Conclusion}
We presented TRACE, an attribute executor for sequential image editing that separates a continuing target from the surface on which it should appear. Active goals retain the user's intended state, local operations recover color on current or acquired pixels, and complete-request checks select the returned result. Matched experiments establish the value of executing attributes on the current surface instead of restoring historical pixels. Evaluations across editors, independent readers, new sources, and complete benchmarks characterize joint recovery and the scene-preservation benefits of local correction. Extending the same target--carrier separation to compound requests and additional appearance attributes is a promising direction.

\clearpage
\section*{LLM Usage Statement}
LLM was used for language polishing. All AI-assisted parts were reviewed and verified by the authors, who take full responsibility for the content of this paper.

{\small\bibliographystyle{ieeenat_fullname}\bibliography{references}}

@inproceedings{lin2014coco,
  title = {{Microsoft COCO: Common Objects in Context}},
  author = {Tsung-Yi Lin and Michael Maire and Serge Belongie and James Hays and Pietro Perona and Deva Ramanan and Piotr Doll{\'a}r and C. Lawrence Zitnick},
  booktitle = {European Conference on Computer Vision},
  pages = {740--755},
  year = {2014},
  doi = {10.1007/978-3-319-10602-1_48},
  url = {https://link.springer.com/chapter/10.1007/978-3-319-10602-1_48},
}

@inproceedings{clip,
  title = {{Learning Transferable Visual Models From Natural Language Supervision}},
  author = {Alec Radford and Jong Wook Kim and Chris Hallacy and Aditya Ramesh and Gabriel Goh and Sandhini Agarwal and Girish Sastry and Amanda Askell and Pamela Mishkin and Jack Clark and Gretchen Krueger and Ilya Sutskever},
  year = {2021},
  booktitle = {International Conference on Machine Learning},
  pages = {8748--8763},
  url = {https://proceedings.mlr.press/v139/radford21a.html}
}

@inproceedings{caron2021dino,
  title = {{Emerging Properties in Self-Supervised Vision Transformers}},
  author = {Caron, Mathilde and Touvron, Hugo and Misra, Ishan and Jégou, Hervé and Mairal, Julien and Bojanowski, Piotr and Joulin, Armand},
  year = {2021},
  url = {https://openaccess.thecvf.com/content/ICCV2021/html/Caron_Emerging_Properties_in_Self-Supervised_Vision_Transformers_ICCV_2021_paper.html},
  booktitle = {Proceedings of the IEEE/CVF International Conference on Computer Vision},
  pages = {9650--9660},
}

@inproceedings{instructpix2pix,
  title = {{InstructPix2Pix: Learning to Follow Image Editing Instructions}},
  author = {Brooks, Tim and Holynski, Aleksander and Efros, Alexei A.},
  year = {2023},
  booktitle = {Proceedings of the IEEE/CVF Conference on Computer Vision and Pattern Recognition},
  pages = {18392--18402},
  url = {https://openaccess.thecvf.com/content/CVPR2023/html/Brooks_InstructPix2Pix_Learning_To_Follow_Image_Editing_Instructions_CVPR_2023_paper.html}
}

@inproceedings{prompt2prompt,
  title = {{Prompt-to-Prompt Image Editing with Cross-Attention Control}},
  author = {Hertz, Amir and Mokady, Ron and Tenenbaum, Jay and Aberman, Kfir and Pritch, Yael and Cohen-Or, Daniel},
  year = {2023},
  url = {https://openreview.net/forum?id=_CDixzkzeyb},
  booktitle = {International Conference on Learning Representations},
}

@inproceedings{sdedit,
  title = {{SDEdit: Guided Image Synthesis and Editing with Stochastic Differential Equations}},
  author = {Meng, Chenlin and He, Yutong and Song, Yang and Song, Jiaming and Wu, Jiajun and Zhu, Jun-Yan and Ermon, Stefano},
  year = {2022},
  booktitle = {International Conference on Learning Representations},
  url = {https://sde-image-editing.github.io/}
}

@article{blendedlatent,
  title = {{Blended Latent Diffusion}},
  author = {Avrahami, Omri and Fried, Ohad and Lischinski, Dani},
  year = {2023},
  journal = {ACM Transactions on Graphics},
  volume = {42},
  number = {4},
  doi = {10.1145/3592450},
  url = {https://omriavrahami.com/blended-latent-diffusion-page/},
  pages = {149:1--149:11},
}

@inproceedings{masactrl,
  title = {{MasaCtrl: Tuning-Free Mutual Self-Attention Control for Consistent Image Synthesis and Editing}},
  author = {Cao, Mingdeng and Wang, Xintao and Qi, Zhongang and Shan, Ying and Qie, Xiaohu and Zheng, Yinqiang},
  year = {2023},
  booktitle = {Proceedings of the IEEE/CVF International Conference on Computer Vision},
  pages = {22560--22570},
  url = {https://openaccess.thecvf.com/content/ICCV2023/html/Cao_MasaCtrl_Tuning-Free_Mutual_Self-Attention_Control_for_Consistent_Image_Synthesis_and_ICCV_2023_paper.html}
}

@inproceedings{diffedit,
  title = {{DiffEdit: Diffusion-based semantic image editing with mask guidance}},
  author = {Couairon, Guillaume and Verbeek, Jakob and Schwenk, Holger and Cord, Matthieu},
  year = {2023},
  booktitle = {International Conference on Learning Representations},
  url = {https://gcouairon.github.io/2022/11/30/diffedit.html}
}

@inproceedings{mgie,
  title = {{Guiding Instruction-based Image Editing via Multimodal Large Language Models}},
  author = {Fu, Tsu-Jui and Hu, Wenze and Du, Xianzhi and Wang, William Yang and Yang, Yinfei and Gan, Zhe},
  year = {2024},
  booktitle = {International Conference on Learning Representations},
  url = {https://proceedings.iclr.cc/paper_files/paper/2024/hash/f0e91b1314fa5eabf1d7ef6d1561ecec-Abstract-Conference.html}
}

@inproceedings{geneval,
  title = {{GenEval: An object-focused framework for evaluating text-to-image alignment}},
  author = {Ghosh, Dhruba and Hajishirzi, Hannaneh and Schmidt, Ludwig},
  year = {2023},
  booktitle = {Advances in Neural Information Processing Systems},
  pages = {52132--52152},
  url = {https://proceedings.neurips.cc/paper_files/paper/2023/hash/a3bf71c7c63f0c3bcb7ff67c67b1e7b1-Abstract-Datasets_and_Benchmarks.html}
}

@inproceedings{tifa,
  title = {{TIFA: Accurate and Interpretable Text-to-Image Faithfulness Evaluation with Question Answering}},
  author = {Hu, Yushi and Liu, Benlin and Kasai, Jungo and Wang, Yizhong and Ostendorf, Mari and Krishna, Ranjay and Smith, Noah A.},
  year = {2023},
  booktitle = {Proceedings of the IEEE/CVF International Conference on Computer Vision},
  pages = {20406--20417},
  url = {https://openaccess.thecvf.com/content/ICCV2023/html/Hu_TIFA_Accurate_and_Interpretable_Text-to-Image_Faithfulness_Evaluation_with_Question_Answering_ICCV_2023_paper.html}
}

@inproceedings{imagic,
  title = {{Imagic: Text-Based Real Image Editing With Diffusion Models}},
  author = {Kawar, Bahjat and Zada, Shiran and Lang, Oran and Tov, Omer and Chang, Huiwen and Dekel, Tali and Mosseri, Inbar and Irani, Michal},
  year = {2023},
  booktitle = {Proceedings of the IEEE/CVF Conference on Computer Vision and Pattern Recognition},
  pages = {6007--6017},
  url = {https://openaccess.thecvf.com/content/CVPR2023/html/Kawar_Imagic_Text-Based_Real_Image_Editing_With_Diffusion_Models_CVPR_2023_paper.html}
}

@inproceedings{nulltext,
  title = {{NULL-Text Inversion for Editing Real Images Using Guided Diffusion Models}},
  author = {Mokady, Ron and Hertz, Amir and Aberman, Kfir and Pritch, Yael and Cohen-Or, Daniel},
  year = {2023},
  booktitle = {Proceedings of the IEEE/CVF Conference on Computer Vision and Pattern Recognition},
  pages = {6038--6047},
  url = {https://openaccess.thecvf.com/content/CVPR2023/html/Mokady_NULL-Text_Inversion_for_Editing_Real_Images_Using_Guided_Diffusion_Models_CVPR_2023_paper.html}
}

@inproceedings{pix2pixzero,
  title = {{Zero-shot Image-to-Image Translation}},
  author = {Parmar, Gaurav and Singh, Krishna Kumar and Zhang, Richard and Li, Yijun and Lu, Jingwan and Zhu, Jun-Yan},
  year = {2023},
  url = {https://graphics.cs.cmu.edu/?p=2636},
  booktitle = {ACM SIGGRAPH Conference Proceedings},
  pages = {1--11},
}

@inproceedings{plugandplay,
  title = {{Plug-and-Play Diffusion Features for Text-Driven Image-to-Image Translation}},
  author = {Tumanyan, Narek and Geyer, Michal and Bagon, Shai and Dekel, Tali},
  year = {2023},
  booktitle = {Proceedings of the IEEE/CVF Conference on Computer Vision and Pattern Recognition},
  pages = {1921--1930},
  url = {https://openaccess.thecvf.com/content/CVPR2023/html/Tumanyan_Plug-and-Play_Diffusion_Features_for_Text-Driven_Image-to-Image_Translation_CVPR_2023_paper.html}
}

@inproceedings{magicbrush,
  title = {{MagicBrush: A Manually Annotated Dataset for Instruction-Guided Image Editing}},
  author = {Zhang, Kai and Mo, Lingbo and Chen, Wenhu and Sun, Huan and Su, Yu},
  year = {2023},
  booktitle = {Advances in Neural Information Processing Systems},
  pages = {31428--31449},
  url = {https://proceedings.neurips.cc/paper_files/paper/2023/hash/64008fa30cba9b4d1ab1bd3bd3d57d61-Abstract.html}
}

@inproceedings{chen2024mimicbrush,
  title = {{Zero-shot Image Editing with Reference Imitation}},
  author = {Chen, Xi and Feng, Yutong and Chen, Mengting and Wang, Yiyang and Zhang, Shilong and Liu, Yu and Shen, Yujun and Zhao, Hengshuang},
  year = {2024},
  url = {https://proceedings.neurips.cc/paper_files/paper/2024/hash/98b2b307aa4aa323df2ba3a83460f25e-Abstract-Conference.html},
  booktitle = {Advances in Neural Information Processing Systems},
  volume = {37},
}

@inproceedings{smartedit,
  title = {{SmartEdit: Exploring Complex Instruction-based Image Editing with Multimodal Large Language Models}},
  author = {Huang, Yuzhou and Xie, Liangbin and Wang, Xintao and Yuan, Ziyang and Cun, Xiaodong and Ge, Yixiao and Zhou, Jiantao and Dong, Chao and Huang, Rui and Zhang, Ruimao and Shan, Ying},
  year = {2024},
  booktitle = {Proceedings of the IEEE/CVF Conference on Computer Vision and Pattern Recognition},
  pages = {8362--8371},
  url = {https://openaccess.thecvf.com/content/CVPR2024/html/Huang_SmartEdit_Exploring_Complex_Instruction-based_Image_Editing_with_Multimodal_Large_Language_CVPR_2024_paper.html}
}

@inproceedings{zone,
  title = {{ZONE: Zero-Shot Instruction-Guided Local Editing}},
  author = {Li, Shanglin and Zeng, Bohan and Feng, Yutang and Gao, Sicheng and Liu, Xiuhui and Liu, Jiaming and Li, Lin and Tang, Xu and Hu, Yao and Liu, Jianzhuang and Zhang, Baochang},
  year = {2024},
  booktitle = {Proceedings of the IEEE/CVF Conference on Computer Vision and Pattern Recognition},
  url = {https://openaccess.thecvf.com/content/CVPR2024/html/Li_ZONE_Zero-Shot_Instruction-Guided_Local_Editing_CVPR_2024_paper.html},
  pages = {6254--6263},
}

@inproceedings{viescore,
  title = {{VIEScore: Towards Explainable Metrics for Conditional Image Synthesis Evaluation}},
  author = {Max Ku and Dongfu Jiang and Cong Wei and Xiang Yue and Wenhu Chen},
  year = {2024},
  booktitle = {Proceedings of the 62nd Annual Meeting of the Association for Computational Linguistics (Volume 1: Long Papers)},
  pages = {12268--12290},
  url = {https://aclanthology.org/2024.acl-long.663/}
}

@inproceedings{sam2,
  title = {{SAM 2: Segment Anything in Images and Videos}},
  author = {Ravi, Nikhila and Gabeur, Valentin and Hu, Yuan-Ting and Hu, Ronghang and Ryali, Chaitanya and Ma, Tengyu and Khedr, Haitham and Rädle, Roman and Rolland, Chloe and Gustafson, Laura and others},
  year = {2025},
  url = {https://proceedings.iclr.cc/paper_files/paper/2025/hash/45c1f6a8cbf2da59ebf2c802b4f742cd-Abstract-Conference.html},
  booktitle = {International Conference on Learning Representations},
}

@inproceedings{rfinversion,
  title = {{Semantic Image Inversion and Editing using Rectified Stochastic Differential Equations}},
  author = {Rout, Litu and Chen, Yujia and Ruiz, Nataniel and Caramanis, Constantine and Shakkottai, Sanjay and Chu, Wen-Sheng},
  year = {2025},
  url = {https://proceedings.iclr.cc/paper_files/paper/2025/hash/c07bc8b520aa87326b015b0da6a70b48-Abstract-Conference.html},
  booktitle = {International Conference on Learning Representations},
}

@inproceedings{emuedit,
  title = {{Emu Edit: Precise Image Editing via Recognition and Generation Tasks}},
  author = {Sheynin, Shelly and Polyak, Adam and Singer, Uriel and Kirstain, Yuval and Zohar, Amit and Ashual, Oron and Parikh, Devi and Taigman, Yaniv},
  year = {2024},
  booktitle = {Proceedings of the IEEE/CVF Conference on Computer Vision and Pattern Recognition},
  pages = {8871--8879},
  url = {https://openaccess.thecvf.com/content/CVPR2024/html/Sheynin_Emu_Edit_Precise_Image_Editing_via_Recognition_and_Generation_Tasks_CVPR_2024_paper.html}
}

@misc{wang2024qwen2vl,
  title = {{Qwen2-VL: Enhancing Vision-Language Model's Perception of the World at Any Resolution}},
  author = {Wang, Peng and Bai, Shuai and Tan, Sinan and Wang, Shijie and Fan, Zhihao and Bai, Jinze and Chen, Keqin and Liu, Xuejing and Wang, Jialin and Ge, Wenbin and others},
  year = {2024},
  howpublished = {arXiv preprint},
  archivePrefix = {arXiv},
  eprint = {2409.12191},
  url = {https://arxiv.org/abs/2409.12191}
}

@misc{qwen3vlreport,
  title = {{Qwen3-VL Technical Report}},
  author = {Bai, Shuai and Cai, Yuxuan and Chen, Ruizhe and Chen, Keqin and Chen, Xionghui and Cheng, Zesen and Deng, Lianghao and Ding, Wei and Gao, Chang and Ge, Chunjiang and others},
  year = {2025},
  howpublished = {arXiv preprint},
  archivePrefix = {arXiv},
  eprint = {2511.21631},
  url = {https://arxiv.org/abs/2511.21631}
}

@misc{edival,
  title = {{EdiVal-Agent: An Object-Centric Framework for Automated, Fine-Grained Evaluation of Multi-Turn Editing}},
  author = {Chen, Tianyu and Zhang, Yasi and Zhang, Zhi and Yu, Peiyu and Wang, Shu and Wang, Zhendong and Lin, Kevin and Wang, Xiaofei and Yang, Zhengyuan and Li, Linjie and others},
  year = {2025},
  howpublished = {arXiv preprint},
  archivePrefix = {arXiv},
  eprint = {2509.13399},
  url = {https://arxiv.org/abs/2509.13399}
}

@inproceedings{flowedit,
  title = {{FlowEdit: Inversion-Free Text-Based Editing Using Pre-Trained Flow Models}},
  author = {Kulikov, Vladimir and Kleiner, Matan and Huberman-Spiegelglas, Inbar and Michaeli, Tomer},
  year = {2025},
  booktitle = {Proceedings of the IEEE/CVF International Conference on Computer Vision},
  pages = {19721--19730},
  url = {https://openaccess.thecvf.com/content/ICCV2025/html/Kulikov_FlowEdit_Inversion-Free_Text-Based_Editing_Using_Pre-Trained_Flow_Models_ICCV_2025_paper.html}
}

@misc{step1xedit,
  title = {{Step1X-Edit: A Practical Framework for General Image Editing}},
  author = {Liu, Shiyu and Han, Yucheng and Xing, Peng and Yin, Fukun and Wang, Rui and Cheng, Wei and Liao, Jiaqi and Wang, Yingming and Fu, Honghao and Han, Chunrui and others},
  year = {2025},
  howpublished = {arXiv preprint},
  archivePrefix = {arXiv},
  eprint = {2504.17761},
  url = {https://arxiv.org/abs/2504.17761}
}

@inproceedings{hpsv3,
  title = {{HPSv3: Towards Wide-Spectrum Human Preference Score}},
  author = {Ma, Yuhang and Wu, Xiaoshi and Sun, Keqiang and Li, Hongsheng},
  year = {2025},
  booktitle = {Proceedings of the IEEE/CVF International Conference on Computer Vision},
  pages = {15086--15095},
  url = {https://openaccess.thecvf.com/content/ICCV2025/html/Ma_HPSv3_Towards_Wide-Spectrum_Human_Preference_Score_ICCV_2025_paper.html}
}

@misc{qwenimage,
  title = {{Qwen-Image Technical Report}},
  author = {Wu, Chenfei and Li, Jiahao and Zhou, Jingren and Lin, Junyang and Gao, Kaiyuan and Yan, Kun and Yin, Sheng-ming and Bai, Shuai and Xu, Xiao and Chen, Yilei and others},
  year = {2025},
  howpublished = {arXiv preprint},
  archivePrefix = {arXiv},
  eprint = {2508.02324},
  url = {https://arxiv.org/abs/2508.02324}
}

@inproceedings{mcedit,
  title = {{Multi-turn Consistent Image Editing}},
  author = {Zhou, Zijun and Deng, Yingying and He, Xiangyu and Dong, Weiming and Tang, Fan},
  year = {2025},
  booktitle = {Proceedings of the IEEE/CVF International Conference on Computer Vision},
  pages = {15792--15801},
  url = {https://openaccess.thecvf.com/content/ICCV2025/html/Zhou_Multi-turn_Consistent_Image_Editing_ICCV_2025_paper.html}
}

@inproceedings{kvedit,
  title = {{KV-Edit: Training-Free Image Editing for Precise Background Preservation}},
  author = {Zhu, Tianrui and Zhang, Shiyi and Shao, Jiawei and Tang, Yansong},
  year = {2025},
  booktitle = {Proceedings of the IEEE/CVF International Conference on Computer Vision},
  pages = {16607--16617},
  url = {https://openaccess.thecvf.com/content/ICCV2025/html/Zhu_KV-Edit_Training-Free_Image_Editing_for_Precise_Background_Preservation_ICCV_2025_paper.html}
}

@misc{fluxkontext,
  title = {{FLUX.1 Kontext: Flow Matching for In-Context Image Generation and Editing in Latent Space}},
  author = {{Black Forest Labs} and Batifol, Stephen and Blattmann, Andreas and Boesel, Frederic and Consul, Saksham and Diagne, Cyril and Dockhorn, Tim and English, Jack and English, Zion and Esser, Patrick and others},
  year = {2025},
  howpublished = {arXiv preprint},
  archivePrefix = {arXiv},
  eprint = {2506.15742},
  url = {https://arxiv.org/abs/2506.15742}
}

@misc{chen2026mtforcing,
  title = {{MT Forcing: Towards Stable and Consistent Multi-Turn Image Editing with Autoregressive Diffusion}},
  author = {Chen, Houyuan and Wang, Fanzhou and Fan, Xiangyu and Qiu, Zesong and Wei, Yukun and Yang, Songlin and Yang, Zhitao and Zhang, Lvmin and Yang, Pei and Zhao, Hao and Lin, Dahua and Yang, Lei and Rao, Anyi},
  year = {2026},
  howpublished = {arXiv preprint},
  archivePrefix = {arXiv},
  eprint = {2610.25404},
  url = {https://arxiv.org/abs/2610.25404}
}

@misc{mtedit,
  title = {{MTEdit: Packed Trajectories Beat Rollouts for Multi-Turn Image Editing}},
  author = {Chen, Houyuan and Wang, Fanzhou and Fan, Xiangyu and Yang, Songlin and Yang, Zhitao and Zhang, Lvmin and Yang, Pei and Zhao, Hao and Lin, Dahua and Yang, Lei and Rao, Anyi},
  year = {2026},
  howpublished = {arXiv preprint},
  archivePrefix = {arXiv},
  eprint = {2610.23751},
  url = {https://arxiv.org/abs/2610.23751}
}

@misc{acedit,
  title = {{ACEdit: Stable Multi-Turn Image Editing via Asymmetric Conditioning}},
  author = {Dorfman, Sara and Goren, Nir and Barak, Gal and Cohen-Or, Daniel and Patashnik, Or},
  year = {2026},
  howpublished = {arXiv preprint},
  archivePrefix = {arXiv},
  eprint = {2610.22328},
  url = {https://arxiv.org/abs/2610.22328}
}

@misc{respec,
  title = {{Making Implicit Preservation Intent Explicit in Conversational Image Editing}},
  author = {Han, Soomin and Ahn, Jihyung and Kim, Bumsoo and Chang, Buru},
  year = {2026},
  howpublished = {arXiv preprint},
  archivePrefix = {arXiv},
  eprint = {2607.07051},
  url = {https://arxiv.org/abs/2607.07051}
}

@misc{mteditflow,
  title = {{MT-EditFlow: Reinforcement Learning for Multi-Turn Image Editing with Flow Matching}},
  author = {Huang, Jiahui and Zhang, Yasi and Chen, Tianyu and Wang, Shu and Xie, Jianwen and Leong, Oscar and Zhou, Mingyuan and Wang, Nanzhu and Wu, Ying Nian},
  year = {2026},
  howpublished = {arXiv preprint},
  archivePrefix = {arXiv},
  eprint = {2606.01985},
  url = {https://arxiv.org/abs/2606.01985}
}

@inproceedings{freqedit,
  title = {{FreqEdit: Preserving High-Frequency Features for Robust Multi-Turn Image Editing}},
  author = {Liao, Yucheng and Liang, Jiajun and Cui, Kaiqian and Zhao, Baoquan and Xie, Haoran and Liu, Wei and Li, Qing and Mao, Xudong},
  year = {2026},
  booktitle = {Proceedings of the IEEE/CVF Conference on Computer Vision and Pattern Recognition},
  pages = {43525--43535},
  url = {https://openaccess.thecvf.com/content/CVPR2026/html/Liao_FreqEdit_Preserving_High-Frequency_Features_for_Robust_Multi-Turn_Image_Editing_CVPR_2026_paper.html}
}

@misc{seam,
  title = {{SEAM: Shot Entity-Attribute Memory for Consistent Short-Drama Generation at Scale}},
  author = {Liu, Jiaqi and Ran, Maolin and Lu, Xiaoyang and Wang, Jian and Liu, Weiwen and Lin, Jianghao and Yu, Yong and Zhang, Weinan},
  year = {2026},
  howpublished = {arXiv preprint},
  archivePrefix = {arXiv},
  eprint = {2608.22725},
  url = {https://arxiv.org/abs/2608.22725}
}

@misc{canvas,
  title = {{CANVAS: Continuity-Aware Narratives via Visual Agentic Storyboarding}},
  author = {Mondal, Ishani and Song, Yiwen and Parmar, Mihir and Goyal, Palash and Boyd-Graber, Jordan and Pfister, Tomas and Song, Yale},
  year = {2026},
  howpublished = {arXiv preprint},
  archivePrefix = {arXiv},
  eprint = {2604.13452},
  url = {https://arxiv.org/abs/2604.13452}
}

@inproceedings{vincie,
  title = {{VINCIE: Unlocking In-context Image Editing from Video}},
  author = {Qu, Leigang and Cheng, Feng and Yang, Ziyan and Zhao, Qi and Lin, Shanchuan and Shi, Yichun and Li, Yicong and Wang, Wenjie and Chua, Tat-Seng and Jiang, Lu},
  year = {2026},
  booktitle = {International Conference on Learning Representations},
  url = {https://proceedings.iclr.cc/paper_files/paper/2026/hash/baf0fab890edc9dce805d7c518058712-Abstract-Conference.html}
}

@misc{overpainting,
  title = {{Overpainting: Localized Context-aware Diffusion Image Editing}},
  author = {Sartor, Sam and Georgiev, Iliyan and Fischer, Michael and Deschaintre, Valentin and Peers, Pieter},
  year = {2026},
  howpublished = {arXiv preprint},
  archivePrefix = {arXiv},
  eprint = {2609.10811},
  url = {https://arxiv.org/abs/2609.10811}
}

@misc{imagagent,
  title = {{IMAGAgent: Orchestrating Multi-Turn Image Editing via Constraint-Aware Planning and Reflection}},
  author = {Shen, Fei and Xie, Chengyu and Wang, Lihong and Zhang, Zhanyi and Jiang, Xin and Du, Xiaoyu and Tang, Jinhui and Chua, Tat-Seng},
  year = {2026},
  howpublished = {arXiv preprint},
  archivePrefix = {arXiv},
  eprint = {2603.29602},
  url = {https://arxiv.org/abs/2603.29602}
}

@inproceedings{wu2026editreward,
  title = {{EditReward: A Human-Aligned Reward Model for Instruction-Guided Image Editing}},
  author = {Wu, Keming and Jiang, Sicong and Ku, Max and Nie, Ping and Liu, Minghao and Chen, Wenhu},
  year = {2026},
  booktitle = {International Conference on Learning Representations},
  url = {https://proceedings.iclr.cc/paper_files/paper/2026/hash/4d68af3056842229f8b356d686aec914-Abstract-Conference.html},
}

@misc{xia2026rrsi,
  title = {{RRSI: Regularized Recursive Self-Improvement of Agent Harnesses}},
  author = {Xia, Peng and Han, Rujun and Wang, Zifeng and Chen, Yanfei and Zhuang, Yufan and Lee, Yoonho and Huang, Chengsong and Yu, Han and CuiZhu, Zhongying and Ming, Yifei and Yao, Huaxiu and Gokturk, Burak and Pfister, Tomas and Lee, Chen-Yu},
  year = {2026},
  howpublished = {arXiv preprint},
  archivePrefix = {arXiv},
  eprint = {2609.24972},
  url = {https://arxiv.org/abs/2609.24972}
}

@misc{xu2026harness2,
  title = {{Harness$^2$: Recursive Agent Harnessing for an Open World}},
  author = {Xu, Ruiyao and Chen, Yanfei and CuiZhu, Zhongying and Mishra, Bhavana Dalvi and Ming, Yifei and Yu, Han and Han, Rujun and Lee, Chen-Yu and Pfister, Tomas},
  year = {2026},
  howpublished = {arXiv preprint},
  archivePrefix = {arXiv},
  eprint = {2610.14224},
  url = {https://arxiv.org/abs/2610.14224}
}

@misc{anchoredit,
  title = {{AnchorEdit: Maintaining Temporal Consistency in Multi-turn Image Editing via Causal Memory}},
  author = {Xu, Hang and Ma, Xiaoxiao and Zhang, Guohui and Hu, Yu and Fu, Siming and Huang, Jie and Song, Lin and Huang, Haoyang and Duan, Nan and Zhao, Feng},
  year = {2026},
  howpublished = {arXiv preprint},
  archivePrefix = {arXiv},
  eprint = {2606.11751},
  url = {https://arxiv.org/abs/2606.11751}
}

@misc{agentbanana,
  title = {{Agent Banana: High-Fidelity Image Editing with Agentic Thinking and Tooling}},
  author = {Ye, Ruijie and Zhang, Jiayi and Liu, Zhuoxin and Zhu, Zihao and Yang, Siyuan and Li, Li and Fu, Tianfu and Dernoncourt, Franck and Zhao, Yue and Zhu, Jiacheng and Rossi, Ryan and Chai, Wenhao and Tu, Zhengzhong},
  year = {2026},
  howpublished = {arXiv preprint},
  archivePrefix = {arXiv},
  eprint = {2602.09084},
  url = {https://arxiv.org/abs/2602.09084}
}

@misc{editr2,
  title = {{Edit-R2: Context-Aware Reinforcement Learning for Multi-Turn Image Editing}},
  author = {Ye, Yuxiao and He, Haoran and Kong, Fangyuan and Wang, Xintao and Wan, Pengfei and Gai, Kun and Pan, Ling},
  year = {2026},
  howpublished = {arXiv preprint},
  archivePrefix = {arXiv},
  eprint = {2606.05950},
  url = {https://arxiv.org/abs/2606.05950}
}

@misc{mtopsd,
  title = {{On-Policy Self-Distillation for Multi-Turn Image Editing}},
  author = {Zhao, Liangbing and Zhuo, Le and Elhoseiny, Mohamed},
  year = {2026},
  howpublished = {arXiv preprint},
  archivePrefix = {arXiv},
  eprint = {2609.35611},
  url = {https://arxiv.org/abs/2609.35611}
}

@misc{flux2klein,
  title = {{FLUX.2 [klein]: Towards Interactive Visual Intelligence}},
  author = {{Black Forest Labs}},
  year = {2026},
  howpublished = {Model release},
  url = {https://bfl.ai/blog/flux2-klein-towards-interactive-visual-intelligence}
}

@misc{simeoni2025dinov3,
  title = {{DINOv3}},
  author = {Sim\'eoni, Oriane and Vo, Huy V. and Seitzer, Maximilian and Baldassarre, Federico and Oquab, Maxime and Jose, Cijo and Khalidov, Vasil and Szafraniec, Marc and Yi, Seungeun and Ramamonjisoa, Micha\"el and others},
  year = {2025},
  eprint = {2508.10104},
  archivePrefix = {arXiv},
  primaryClass = {cs.CV},
  url = {https://arxiv.org/abs/2508.10104}
}

@inproceedings{liu2023groundingdino,
  title = {{Grounding DINO: Marrying DINO with Grounded Pre-Training for Open-Set Object Detection}},
  author = {Liu, Shilong and Zeng, Zhaoyang and Ren, Tianhe and Li, Feng and Zhang, Hao and Yang, Jie and Jiang, Qing and Li, Chunyuan and Yang, Jianwei and Su, Hang and Zhu, Jun and Zhang, Lei},
  year = {2024},
  url = {https://www.ecva.net/papers/eccv_2024/papers_ECCV/html/6319_ECCV_2024_paper.php},
  booktitle = {European Conference on Computer Vision},
  pages = {38--55},
  doi = {10.1007/978-3-031-72970-6_3},
}

@misc{huang2026editjudgebias,
  title = {{Do MLLM Judges Judge the Edit? Auditing Bias in Image Editing Evaluation with Verified Quality Preservation}},
  author = {Huang, Yuan and Song, Zirui and Chen, Xiuying},
  year = {2026},
  eprint = {2610.01670},
  archivePrefix = {arXiv},
  primaryClass = {cs.CV},
  url = {https://arxiv.org/abs/2610.01670}
}

@article{reinhard2001color,
  author = {Erik Reinhard and Michael Ashikhmin and Bruce Gooch and Peter Shirley},
  title = {Color Transfer between Images},
  journal = {IEEE Computer Graphics and Applications},
  volume = {21},
  number = {5},
  pages = {34--41},
  year = {2001},
  doi = {10.1109/38.946629}
}

@misc{ottosson2020oklab,
  author = {Bj{\"o}rn Ottosson},
  title = {A Perceptual Color Space for Image Processing},
  year = {2020},
  howpublished = {\url{https://bottosson.github.io/posts/oklab/}}
}

@inproceedings{gupta2023visual,
  author = {Tanmay Gupta and Aniruddha Kembhavi},
  title = {Visual Programming: Compositional Visual Reasoning Without Training},
  booktitle = {Proceedings of the IEEE/CVF Conference on Computer Vision and Pattern Recognition},
  pages = {14953--14962},
  year = {2023},
  url = {https://openaccess.thecvf.com/content/CVPR2023/html/Gupta_Visual_Programming_Compositional_Visual_Reasoning_Without_Training_CVPR_2023_paper.html}
}

@article{wang2023voyager,
  author = {Guanzhi Wang and Yuqi Xie and Yunfan Jiang and Ajay Mandlekar and Chaowei Xiao and Yuke Zhu and Linxi Fan and Anima Anandkumar},
  title = {{Voyager}: An Open-Ended Embodied Agent with Large Language Models},
  journal = {Transactions on Machine Learning Research},
  year = {2024},
  url = {https://openreview.net/forum?id=ehfRiF0R3a},
}

@inproceedings{colorctrl,
  author = {Yin, Zixin and Dai, Xili and Chen, Ling-Hao and Zhou, Deyu and Wang, Jianan and Wang, Duomin and Yu, Gang and Ni, Lionel M. and Zhang, Lei and Shum, Heung-Yeung},
  title = {Training-Free Text-Guided Color Editing with Multi-Modal Diffusion Transformer},
  booktitle = {International Conference on Learning Representations},
  year = {2026},
  url = {https://proceedings.iclr.cc/paper_files/paper/2026/hash/59f278de1619bdb6b53fd04e8e0976e0-Abstract-Conference.html},
}

@misc{coloredit,
  author = {Yin, Xingxi and Li, Zhi and Zhang, Jingfeng and Li, Chenglin and Zhang, Yin},
  title = {{ColorEdit}: Training-free Image-Guided Color editing with diffusion model},
  year = {2024},
  archivePrefix = {arXiv},
  eprint = {2411.10232},
  url = {https://arxiv.org/abs/2411.10232v1}
}

@inproceedings{kim2025layerwisememory,
  author = {Daneul Kim and Jaeah Lee and Jaesik Park},
  title = {Improving Editability in Image Generation with Layer-wise Memory},
  booktitle = {Proceedings of the IEEE/CVF Conference on Computer Vision and Pattern Recognition},
  year = {2025},
  url = {https://openaccess.thecvf.com/content/CVPR2025/html/Kim_Improving_Editability_in_Image_Generation_with_Layer-wise_Memory_CVPR_2025_paper.html},
  pages = {7889--7898}
}

@misc{glm2025thinking,
  author = {{GLM-V Team}},
  title = {{GLM-4.5V} and {GLM-4.1V-Thinking}: Towards Versatile Multimodal Reasoning with Scalable Reinforcement Learning},
  year = {2025},
  archivePrefix = {arXiv},
  eprint = {2507.01006},
  url = {https://arxiv.org/abs/2507.01006}
}

@misc{yang2026rise,
  author = {Yang, Xue and Zhang, Peiyuan and Zhu, Yilun and Yang, Qihao and Liu, Mingxin and Zhao, Xiangyu and Fan, Ziqian and Wang, Zhaokai and Li, Yan and Yang, Yifan and Yang, Xu and Jia, Xiaosong and Zhou, Yue and Zhong, Zhihang and Yan, Junchi},
  title = {Reasoning-Informed Visual Editing},
  year = {2026},
  archivePrefix = {arXiv},
  eprint = {2610.12343},
  url = {https://arxiv.org/abs/2610.12343}
}

@inproceedings{qu2026replan,
  author = {Qu, Tianyuan and Ke, Lei and Zhan, Xiaohang and Tang, Longxiang and Liu, Yuqi and Peng, Bohao and Yu, Bei and Yu, Dong and Jia, Jiaya},
  title = {{RePlan}: Reasoning-guided Region Planning for Complex Instruction-based Image Editing},
  booktitle = {European Conference on Computer Vision},
  year = {2026},
  pages = {616--631},
  doi = {10.1007/978-3-032-37718-0_35},
  url = {https://link.springer.com/chapter/10.1007/978-3-032-37718-0_35}
}

@article{robertson2009bm25,
  title = {{The Probabilistic Relevance Framework: BM25 and Beyond}},
  author = {Stephen Robertson and Hugo Zaragoza},
  journal = {Foundations and Trends in Information Retrieval},
  volume = {3},
  number = {4},
  pages = {333--389},
  year = {2009},
  doi = {10.1561/1500000019},
  url = {https://www.nowpublishers.com/article/Details/INR-019},
}

@misc{guan2026bluepixel,
  author = {Shanyan Guan and Shiyu Li and Xin Li and Zhizhou Chen and Yunzhe Xu and Yuehao Liu and Ruixue Shen and Shiju Zhang and Min Jin and Qiang Liu and others},
  title = {{BluePixel}: Toward Native {4K} Agentic Image Editing},
  year = {2026},
  archivePrefix = {arXiv},
  eprint = {2610.19882},
  url = {https://arxiv.org/abs/2610.19882}
}

\clearpage
% Supplementary material, with its original text and numbering.
\setcounter{dbltopnumber}{3}
\setcounter{equation}{0}
\setcounter{figure}{0}
\setcounter{table}{0}
\setcounter{algorithm}{0}
\setcounter{footnote}{0}
% Use distinct hyperlink destinations after resetting the counters.
\renewcommand{\theHequation}{supp.\arabic{equation}}
\renewcommand{\theHfigure}{supp.\arabic{figure}}
\renewcommand{\theHtable}{supp.\arabic{table}}
\providecommand{\theHalgorithm}{\arabic{algorithm}}
\renewcommand{\theHalgorithm}{supp.\arabic{algorithm}}
\TRACErestoresupplementtitle
\maketitle\appendix\renewcommand{\theequation}{S\arabic{equation}}\renewcommand{\thefigure}{S\arabic{figure}}\renewcommand{\thetable}{S\arabic{table}}
\renewcommand{\thealgorithm}{S\arabic{algorithm}}
\noindent We describe the execution rules and their conditional properties (Secs.~\ref{app:execution}--\ref{app:interval-certificate}), evaluation protocols (Sec.~\ref{app:protocols}), and complete benchmark and controlled results (Secs.~\ref{app:legacy-results}--\ref{app:material-recovery}).

\section{Attribute execution on current samples}
\label{app:execution}
\label{app:program-properties}
Let $A$ be the current proposal after optional spatial restriction. The carrier $D$ is either $A$ or a corrective image $R$, with Oklab samples $x_p=(L_p,a_p,b_p)=\phi(D(p))$ on a nonempty support $M$. Here $\phi$ converts encoded RGB to Oklab, $m$ is the coordinatewise sample median, and qualified evidence supplies target center $\theta$. For even sample counts, the median averages the two central values. We assume $L_p,\theta_L\in[0,1]$. Sample-space identities precede gamut mapping and RGB encoding unless stated otherwise.

\paragraph{Binding the target and carrier.} Complete introducing requests and intervening instructions determine goal validity, including entity identity and membership. Qualified history supplies the color center; current pixels supply the surface. Carrier acquisition applies to one supported atomic material request and one continuing color on the same entity. An observer receives only a neutral category and image; a known material mismatch invokes acquisition. Other cases follow the color-recovery route in Algorithm~\ref{alg:local-execution}.

\begin{algorithm}[t]
\caption{Local color-recovery route in TRACE}
\label{alg:local-execution}
\small
\begin{algorithmic}[1]
\Require Current proposal $A$, active goals $\mathcal G$, returned history $\mathcal H$, retained operations $\mathcal P$
\Ensure Checked local update, or unchanged proposal $A$
\State $g\gets$ first eligible continuing color goal with a known mismatch
\If{$g$ is unavailable} \State \Return $A$ \EndIf
\State $M\gets\Call{UniqueSupport}{A,g}$
\If{$M$ is unavailable} \State \Return $A$ \EndIf
\If{$g$ requests gray}
    \State $C\gets\Call{ZeroChroma}{A,M}$
    \If{$\Call{Check}{C,A,g,\mathcal G}$} \State \Return $C$ \EndIf
\EndIf
\State $(f_1,\ldots,f_k)\gets\Call{FamilyOrder}{g}$
\For{$f\in(f_1,\ldots,f_k)$}
    \For{$s\in(\mathsf{Retained},\mathsf{LatestHistory})$}
        \State $\theta\gets\Call{Parameters}{s,g,f,\mathcal P,\mathcal H}$
        \If{$\theta$ is available}
            \State $C\gets T^f_\theta(A;M)$, copying $A$ outside $M$
            \If{$\Call{Check}{C,A,g,\mathcal G}$} \State \Return $C$ \EndIf
        \EndIf
    \EndFor
\EndFor
\State $R\gets\Call{CorrectiveImage}{A,\mathcal G}$ \Comment{at most one draw}
\If{$R$ is unavailable} \State \Return $A$ \EndIf
\State $\theta\gets\Call{FitCenter}{R|_M}$
\State $C_{\mathrm{op}}\gets T^{f_1}_\theta(A;M)$
\State $C_{\mathrm{copy}}\gets R$ on $M$, $A$ elsewhere
\State $\mathcal F\gets\{C\in\{C_{\mathrm{op}},C_{\mathrm{copy}}\}:\Call{Check}{C,A,g,\mathcal G}\}$
\If{$\mathcal F=\varnothing$} \State \Return $A$ \EndIf
\State \Return $\Call{LexicographicBest}{\mathcal F;-N_T,E_L,D_0}$
\end{algorithmic}
\end{algorithm}

Within each family, retained parameters precede the latest qualified history; all families use the same historical support. Every \textsc{Check} requires the target color, its complete introducing request and the current request to pass, with interval nondecrease for other checked requests. Reuse rebinds goals and supports; tonal refinement is tested once after parent selection.

\paragraph{Selecting a local correction.} For encoded images $C$ and $A$, let $N_T(C)$ count positively checked requests and $D_0(C,A)=\sum_p\mathbf{1}[C(p)\ne A(p)]$ count changed pixels. Let $L_p^C$ be the decoded Oklab lightness of $C$, and let $\overline L^C$ be its mean over $M$. The centered-lightness objective is
\begin{equation}
E_L(C,A)=\frac{1}{|M|}\sum_{p\in M}\left|(L_p^C-\overline L^C)-(L_p^A-\overline L^A)\right|.
\label{eq:lightness-objective}
\end{equation}
For the finite set $\mathcal F_g^R(A)$ of feasible ordinary color-recovery candidates from the single corrective draw, selection is
\begin{equation}
\widehat C=\underset{C\in\mathcal F_g^R(A)}{\arg\min}\;\bigl(-N_T(C),E_L(C,A),D_0(C,A)\bigr),
\label{eq:finite-selection}
\end{equation}
The minimum is lexicographic: more satisfied requests precede less lightness disturbance and then fewer changed pixels. Ties favor the parametric operation; an empty set retains $A$. Carrier acquisition uses the selection rule below.

\paragraph{Acquiring a local carrier.} For a known material mismatch, color recovery without generation supplies fallback $P$. Generate $R$ from $A$ using the current request, the neutral category's continuing color, and material-specific texture and reflectance cues. The current request takes precedence; the instruction also asks to preserve shape, pose, markings, position, lighting and unrelated content. The rendering instruction names the neutral category and desired attributes; admission checks complete introducing requests against their original input images. Fresh support $M$ is bound on $A$ and used as a fixed window on $R$. The two candidate types are
\begin{equation}
\begin{aligned}
C_{\rm copy}(p)&=\begin{cases}R(p),&p\in M,\\A(p),&p\notin M,\end{cases}\\
C_f(p)&=\begin{cases}\mathsf{Encode}(T^f_\theta(\phi(R(p)))),&p\in M,\\A(p),&p\notin M.\end{cases}
\end{aligned}
\label{eq:carrier-candidates}
\end{equation}
Qualified history supplies $\theta$; each family recenters on $R|_M$, and encoding includes radial gamut mapping. Chromatic goals use chroma and full families; black, white, and gray use the full family. Feasible candidates pass fresh color, complete introducing-request and current-request checks; other earlier requests are compared with $P$. Rank feasible candidates by positive neutral material, $E_L(C,R)$, $D_0(C,A)$, then local copying on ties. The material reading guides ranking among these feasible candidates. Return $P$ if none passes. Both routes share one draw per proposal. Acquisition preserves parameter memory; only separately checked color-recovery operations are retained.

\paragraph{Refining the selected operation.} A selected chroma/full operation within the scope below receives one tonal refinement using the same carrier, target, support and encoding. Fresh color and complete-request checks determine replacement; rejection retains the parent. The carrier, draw count and parameter memory remain fixed.

\paragraph{Checked global fallback.} Let $C$ be the selected and refined local result, with $B=A$ for color recovery or $B=P$ for acquisition. Return existing $R$ only if encoded $C=B=A$ and $R$ passes neutral color and every complete active request; otherwise retain $C$. This preserves accepted local progress within a proposal, with the same masks and memory. Exterior-pixel identity applies to local candidates.

\paragraph{A second material proposal.} For qualified atomic wood/metal acquisition with a known mismatch on $A$, let $C_1$ be the complete first-proposal result. If either neutral color or material is not $T$, one additional draw uses a fixed physical-finish instruction and the same $A$, target, support and sampler seed. The same local executor, calibration and strict fallback produce $C_2$. Replace $C_1$ only if $C_2$ passes both neutral attributes and all complete active requests; otherwise retain $C_1$. This route permits two corrective draws; other routes permit one.

\paragraph{Bounded lightness.} The full operation is
\begin{equation}
T_\theta(x_p)=\bigl(\theta_L+s(L_p-m_L),\ a_p+\theta_a-m_a,\ b_p+\theta_b-m_b\bigr).
\label{eq:program-transform}
\end{equation}
Write $L_-=\min_pL_p$ and $L_+=\max_pL_p$. The transformed extrema must satisfy
\begin{align}
\theta_L-s(m_L-L_-)&\geq0,\label{eq:lower-scale-bound}\\
\theta_L+s(L_+-m_L)&\leq1.\label{eq:upper-scale-bound}
\end{align}
If $m_L>L_-$, Eq.~\ref{eq:lower-scale-bound} gives $s\leq\theta_L/(m_L-L_-)$. Otherwise its left side is $\theta_L\geq0$ and imposes no bound. Similarly, Eq.~\ref{eq:upper-scale-bound} gives $s\leq(1-\theta_L)/(L_+-m_L)$ when $L_+>m_L$ and is automatic otherwise. Define the corresponding inactive bounds as $+\infty$. The maximal feasible nonamplifying scale is therefore
\begin{equation}
s^*=\min\left\{1,\frac{\theta_L}{m_L-L_-},\frac{1-\theta_L}{L_+-m_L}\right\},
\label{eq:program-scale}
\end{equation}
with those inactive-bound conventions. Every term is nonnegative. Substituting $s^*$ satisfies both inequalities, and any feasible $s\in[0,1]$ obeys all three bounds, proving maximality within this family.

\paragraph{Center and contrast.} For ordered scalar samples, a nonnegative affine map preserves their order, including the constant case. Applying the map to the central sample, or averaging its two central samples, gives $\operatorname{med}(az+b)=a\operatorname{med}(z)+b$. Applying this identity to the three channels yields $\operatorname{med}_pT_\theta(x_p)=\theta$. Translations also cancel in pairwise differences, so
\begin{align}
\|T_\theta(x_p)-T_\theta(x_q)\|_2^2
&=(s^*)^2(L_p-L_q)^2+(a_p-a_q)^2\notag\\
&\quad+(b_p-b_q)^2\leq\|x_p-x_q\|_2^2.
\end{align}
For $\overline L=|M|^{-1}\sum_pL_p$, averaging the lightness transform gives $\overline L'=\theta_L+s(\overline L-m_L)$. Subtracting it from each transformed sample and then subtracting the original centered sample gives
\begin{align}
(L'_p-\overline L')-(L_p-\overline L)&=-(1-s)(L_p-\overline L),\\
E_L&=\frac{1-s}{|M|}\sum_p|L_p-\overline L|.
\label{eq:lightness-cost}
\end{align}
The sum is fixed and nonnegative, so $s^*$ minimizes this centered-lightness distortion over the feasible family. Actual candidate ranking recomputes $E_L$ on decoded RGB, accounting for gamut mapping and quantization.

\paragraph{Chroma, gray, and gamut.} Chroma execution adds the constant $d=(0,\theta_a-m_a,\theta_b-m_b)$ to each $x_p$. Hence $L'_p=L_p$, $T(x_p)-T(x_q)=x_p-x_q$, and its median is $(m_L,\theta_a,\theta_b)$. Chromatic targets try this family before the full operation. Black and white use the full family; gray first tries $(L,0,0)$. The exact neutral inverse gives equal linear RGB channels $L^3$, and equal transfer functions and rounding preserve their equality.

For other transformed colors, radial mapping tests $(L,\lambda a,\lambda b)$, $\lambda\in[0,1]$. At $\lambda=0$, the neutral inverse gives $(L^3,L^3,L^3)\in[0,1]^3$. Starting from this feasible endpoint, eighteen midpoint tests update the retained endpoint only when RGB is feasible. Induction preserves feasibility, with tolerance $10^{-7}$ and clipping during encoding; a numerical failure rejects the candidate. The guarantee is feasibility of the retained endpoint, including disconnected ray intersections. Ranking recomputes lightness after encoding, while complete-request and appearance checks assess the visible material.

\paragraph{Calibrated tapered chroma.} Index the $n=|M|$ samples of the selected carrier by $i$, and require $0<\theta_L<1$ and $\mu=\operatorname{med}_iL_i\in(0,1)$. Set $w_i=L_i/\mu$ for $L_i\leq\mu$ and $w_i=(1-L_i)/(1-\mu)$ otherwise. These weights lie in $[0,1]$ and remain fixed during fitting. For one chroma coordinate $c\in\{a,b\}$, define $g(d)=\operatorname{med}_i(c_i+w_i d)$. Each sample is continuous and nondecreasing in $d$. Writing $z_i(d)=c_i+w_i d$, the $k$th smallest sample has the finite min--max representation
\begin{equation}
z_{(k)}(d)=\min_{S\subseteq\{1,\ldots,n\}:\,|S|=k}\;\max_{i\in S}z_i(d).
\end{equation}
Finite minima and maxima preserve continuity and monotonicity. The central order statistic, or the average of the central pair, is therefore continuous and nondecreasing. Assume that strictly more than half the weights are positive. As $d\to+\infty$, that majority eventually exceeds every fixed zero-weight sample, placing the central sample or pair in the diverging majority; the argument reverses for $d\to-\infty$. Consequently
\begin{equation}
\lim_{d\to-\infty}g(d)=-\infty,\qquad\lim_{d\to+\infty}g(d)=+\infty.
\end{equation}
Continuity gives a root of $g(d)=\theta_c$, and monotonicity makes its root set a bounded closed interval $[d_-,d_+]$. Its unique point nearest zero is
\begin{equation}
d_c=\begin{cases}0,&g(0)=\theta_c,\\\inf\{d:g(d)\geq\theta_c\},&g(0)<\theta_c,\\\sup\{d:g(d)\leq\theta_c\},&g(0)>\theta_c.\end{cases}
\label{eq:app-calibrated-root}
\end{equation}
This convention handles plateaus: an arbitrary root can have unnecessarily large displacement. The two coordinates have independent root intervals. Their squared sample displacement is
\begin{equation}
\sum_i\|w_i d\|_2^2=\left(\sum_iw_i^2\right)(d_a^2+d_b^2).
\end{equation}
The positive factor is fixed, so the two nearest-zero roots jointly minimize this displacement among all shifts attaining both target medians within the fixed tapered family.

\paragraph{A finite root bracket.} Let $\eta=\min_{i:w_i>0}w_i$, and let $c_-^+,c_+^+$ be the extrema of the corresponding chroma samples. Define
\begin{equation}
d_{\rm lo}=\min\{0,(\theta_c-c_+^+)/\eta\},\quad d_{\rm hi}=\max\{0,(\theta_c-c_-^+)/\eta\}.
\end{equation}
For $d_{\rm lo}\leq0$ and $w_i\geq\eta$, $c_i+w_i d_{\rm lo}\leq c_+^++\eta d_{\rm lo}\leq\theta_c$ on the positive-weight majority. Thus $g(d_{\rm lo})\leq\theta_c$. Similarly, $c_i+w_i d_{\rm hi}\geq c_-^++\eta d_{\rm hi}\geq\theta_c$, giving $g(d_{\rm hi})\geq\theta_c$. If $g(0)<\theta_c$, bisection on $[0,d_{\rm hi}]$ moves the upper bound whenever $g(d)\geq\theta_c$; if $g(0)>\theta_c$, bisection on $[d_{\rm lo},0]$ moves the lower bound whenever $g(d)\leq\theta_c$. These one-sided predicates converge to the nearest plateau boundary. Numerically, the bracket is rounded outward and bisected to adjacent floating-point values, followed by a median-residual check. Refinement requires the positive-weight majority, a finite bracket, convergence and an acceptable residual.

\paragraph{Exact lightness-median calibration.} A full refinement seeks an interior anchor $0<\tau<1$ and uses
\begin{equation}
f_{\mu,\tau}(L)=\begin{cases}\tau L/\mu,&L\leq \mu,\\\tau+(1-\tau)(L-\mu)/(1-\mu),&L>\mu.\end{cases}
\label{eq:app-tone-map}
\end{equation}
An odd mask has central sample $\mu$, so $\tau=\theta_L$. For an even mask, write its central pair as $\mu-q/2,\mu+q/2$. Monotonicity retains that pair as the central pair after transformation. Substitution gives
\begin{align}
f(\mu-q/2)&=\tau\left(1-\frac{q}{2\mu}\right),\\
f(\mu+q/2)&=\tau+(1-\tau)\frac{q}{2(1-\mu)},\\
\operatorname{med}_i f(L_i)&=u+(1-u-v)\tau,
\end{align}
where $u=q/[4(1-\mu)]$ and $v=q/(4\mu)$. The required anchor is therefore
\begin{equation}
\tau=\frac{\theta_L-u}{1-u-v}.
\end{equation}
Because $q\leq2\min(\mu,1-\mu)$,
\begin{equation}
u+v=\frac{q}{4\mu(1-\mu)}\leq\frac{1}{2\max(\mu,1-\mu)}\leq1.
\end{equation}
Hence $1-u-v\geq0$. Equality is possible only for the central pair $(0,1)$ and $\mu=1/2$, when the median is fixed at $1/2$ for every anchor; refinement is withheld. Otherwise strict monotonicity requires $0<\tau<1$, equivalently $u<\theta_L<1-v$. An infeasible anchor is withheld rather than clipped. The implementation evaluates the two central offsets separately to account for floating-point rounding of $\mu$, and verifies the transformed median.

\paragraph{Tonal structure and scope.} The map fixes zero and one. For $\alpha=\tau/\mu>0$, $\beta=(1-\tau)/(1-\mu)>0$, and $L_i\geq L_j$, each segment contributes its length times one of these slopes, yielding
\begin{equation}
\min(\alpha,\beta)(L_i-L_j)\leq f(L_i)-f(L_j)\leq\max(\alpha,\beta)(L_i-L_j).
\end{equation}
Tonal order is preserved, with a finite bound on contrast change. Also $w_\mu$ is $1/\min(\mu,1-\mu)$-Lipschitz and has range $[0,1]$, so the change in a chroma difference obeys
\begin{equation}
\|(c'_i-c'_j)-(c_i-c_j)\|_2\leq\|d\|_2\min\left\{1,\frac{|L_i-L_j|}{\min(\mu,1-\mu)}\right\}.
\end{equation}
Here $c_i=(a_i,b_i)$. These sample-space properties precede gamut mapping and encoding; image-based checks assess visible color, material and complete-request satisfaction. Refinement applies to a selected chroma or full operation only when its complete applicable requests are one source-qualified continuing color request and an atomic current material setter on the same object. Copies, analytic gray and cases outside this scope retain the selected output, as do numerical failures. The carrier, evidence binding and parameter memory remain fixed.

\begin{figure*}[t]
\centering
\includegraphics[width=\textwidth]{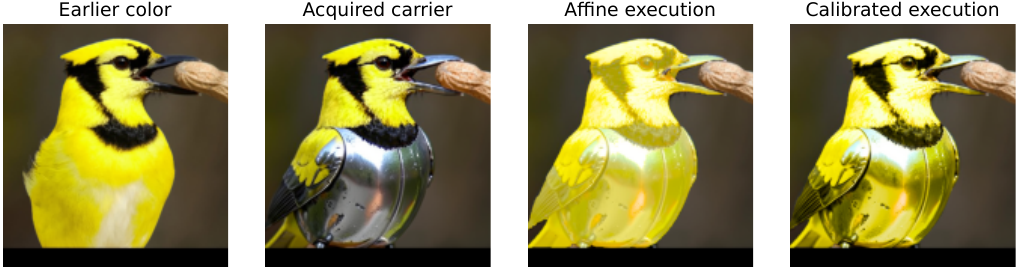}
\caption{Fixed-carrier calibration control after requests for yellow, then metal, with identical crops. The carrier has reflectance but incomplete yellow; affine execution brightens dark markings. Calibration uses the same carrier, target and mask to recover yellow while retaining markings and reflection contrast.}
\label{fig:calibration-case}
\end{figure*}

\paragraph{Half-step control.} On atomic same-object metal requests, repeat $\vartheta^{k+1}=m^{ab}+\tfrac12(\vartheta^k-m^{ab})$ with $A,M$ fixed. Fresh color and complete-request checks admit each recomputed candidate; failure or unavailability stops the control. TRACE uses zero half-steps by default.

Every local operation and corrective-region copy writes only on $M$. Keeping $A$ writes nothing. Therefore any selected local candidate satisfies, after integer encoding,
\begin{equation}
\{p:C(p)\neq A(p)\}\subseteq M,\qquad D_0(C,A)\leq|M|.
\end{equation}
A corrective rendering supplies samples for the local operation; its unselected pixels do not enter the returned image.

\section{Goal continuity, locality, and checked progress}
\label{app:interval-certificate}
The following properties distinguish changes to requested values, changes to pixels, and changes to complete-request satisfaction. Semantic nonregression and progress are conditional on correct bindings and sound visual judgments.
\paragraph{Requested-state continuity.} Let $\mathcal A_t$ be newly introduced goals and $\mathcal R_t$ the user-authorized retirements. The active set evolves as
\begin{equation}
\mathcal G_t=(\mathcal G_{t-1}\setminus\mathcal R_t)\cup\mathcal A_t.
\label{eq:app-live-ledger}
\end{equation}
Let $F$ update requested state $S_t$ from instruction $u_t$ and admitted interpretation $q_t$. Two runs with the same $S_0,u_t,q_t$ satisfy $S_t=F(S_{t-1},u_t,q_t)=F(S'_{t-1},u_t,q_t)=S'_t$ by induction. The requested state therefore evolves with the instructions and their interpretations. With entity identity, membership and applicability fixed, let $U_{e,b,w}$ update only $V(e,b)$ to $w$ in requested values $V$, and let $\pi_{e,a}$ read entry $(e,a)$. Distinct-attribute setters modify separate entries; for $a\neq b$,
\begin{equation}
\pi_{e,a}\bigl(U_{e,b,w}(V)\bigr)=V(e,a).
\label{eq:source-commutation}
\end{equation}
Repeated complementary attribute setters preserve this equality by induction. Same-slot revision retires the old value; deletion, replacement, and membership changes use their corresponding lifetime rules. Goal binding checks the complete introducing request, intervening instructions, aliases, identity and membership before carrying an appearance goal forward. Complete-request selection separately assesses the resulting images.

\paragraph{Spatial restriction.}
\label{app:commit-properties}
For aligned input $B$, ordinary proposal $Y$, and admitted region $W$, let $Z=Y$ on $W$ and $Z=B$ elsewhere. With $\Delta(C,B)=\{p:C(p)\neq B(p)\}$, substitution gives
\begin{equation}
\Delta(Z,B)=W\cap\Delta(Y,B)\subseteq\Delta(Y,B).
\label{eq:changed-set-inclusion}
\end{equation}
For $D(C,B)=\sum_p d_p(C(p),B(p))$ with $d_p\geq0$ and $d_p(B(p),B(p))=0$, substitution gives $D(Z,B)=\sum_{p\in W}d_p(Y(p),B(p))\leq D(Y,B)$. If a property depends only on a complete support $Q\subseteq W$, agreement on $Q$ preserves it; predicted object masks alone do not establish complete semantic support. For nonempty masks $M_B,M_Y$ with intersection size $c$, directional coverage $c/|M_B|,c/|M_Y|\geq\eta\in(0,1]$ gives $|M_B|,|M_Y|\leq c/\eta$ and therefore $\operatorname{IoU}\geq c/(2c/\eta-c)=\eta/(2-\eta)$. Coverage is checked separately for every plausible target instance.

\paragraph{Local admission and progress.}
\label{app:certificate-first}
\label{app:progress-certificate}
Fix the checked set $\mathcal E$ of complete earlier requests. For each $g\in\mathcal E$, let $r_g(I)\in\{0,1\}$ be its truth under the original introducing input, and let $r_c(I)$ be the truth of the complete current request. Define $\Phi(I)=r_c(I)+\sum_{g\in\mathcal E}r_g(I)$. An observed judgment $v_g(I)\in\{T,F,U\}$ determines an interval $J(v_g(I))$ of possible binary truths, with lower and upper endpoints $\ell(v)$ and $u(v)$.
\begin{equation}
J(F)=[0,0],\qquad J(U)=[0,1],\qquad J(T)=[1,1].
\label{eq:observation-intervals}
\end{equation}
If these intervals contain the corresponding complete-request truths and $\ell(v_g(C))\geq u(v_g(A))$, then
\begin{equation}
r_g(C)\geq\ell(v_g(C))\geq u(v_g(A))\geq r_g(A).
\label{eq:interval-nonregression}
\end{equation}
Spatial selection additionally requires more positive judgments, or identical answers with fewer changed pixels. An all-$T$ restricted candidate with $D_0(Z,B)<D_0(Y,B)$ settles this comparison without querying $Y$: every request is true on $Z$, and either $Y$ has fewer positive judgments or the changed-pixel comparison resolves the tie. A hard restriction differing from $Y$ removes at least one changed pixel, so its inequality is strict. Unavailable judgments remain unresolved. For a removal, the neutral source class is checked on the input; restriction requires ruling out a continuing instance.

For local color repair, let $g_\star$ be its complete earlier request and $f_\star$ its required binary color predicate. Necessity gives $r_{g_\star}(A)\leq f_\star(A)$. A sound known color failure gives $f_\star(A)=0$, so $r_{g_\star}(A)=0$. Sound positive checks give $r_{g_\star}(C)=r_c(C)=1$. For another earlier request with candidate judgment $T$, $r_g(C)=1\geq r_g(A)$ directly, without querying $A$. Otherwise the intervals for $A$ and $C$ give nondecrease by Eq.~\ref{eq:interval-nonregression}. Summing gives
\begin{align}
\Phi(C)-\Phi(A)
&=1+1-r_c(A)+\sum_{g\neq g_\star}\bigl(r_g(C)-r_g(A)\bigr)\notag\\
&\geq1.
\end{align}
Refinement from parent $C_k$ to $C_{k+1}$ retains both positive requests and repeats the remaining comparisons, hence $\Phi(C_{k+1})\geq\Phi(C_k)\geq\Phi(A)+1$. The progress bound applies under fixed checked requests and source bindings.

\paragraph{Progress after carrier acquisition.} Let $g_\star\in\mathcal E$ be the continuing color request, and $f_m(I)$ the necessary material predicate of the complete current request. Correct binding and a sound triggering failure give
\begin{equation}
r_c(I)\leq f_m(I),\qquad f_m(A)=0\ \Longrightarrow\ r_c(A)=0.
\end{equation}
Sound admission gives $r_c(C)=r_{g_\star}(C)=1$. The zero-draw fallback $P$ is $A$ or an update admitted by the color-recovery rule above, which preserves every checked earlier request. Acquisition compares the other requests with $P$. Thus for $g\in\mathcal E\setminus\{g_\star\}$,
\begin{align}
r_g(P)&\geq r_g(A),\\
r_g(C)&\geq\ell(v_g(C))\geq u(v_g(P))\notag\\
&\geq r_g(P)\geq r_g(A).
\end{align}
Since $r_{g_\star}(A)\in\{0,1\}$, summing gives
\begin{align}
\Phi(C)-\Phi(A)
&=1+\bigl[1-r_{g_\star}(A)\bigr]\notag\\
&\quad+\sum_{g\neq g_\star}\bigl[r_g(C)-r_g(A)\bigr]\geq1.
\end{align}
This establishes progress even when the earlier color is already correct on $A$. Admission establishes the bound; neutral material readings rank the feasible candidates.

\paragraph{Progress after global fallback.} A corrective draw is triggered by a known continuing-color or current-material failure. Under sound judgments, at least one checked request is false on $A$, so $\Phi(A)\leq|\mathcal E|$. Global admission makes every checked request true on $R$, yielding $\Phi(R)=|\mathcal E|+1$ and $\Phi(R)-\Phi(A)\geq1$. An admitted $C_2$ has the same all-positive condition, so $\Phi(C_2)\geq\Phi(C_1)$. The bound concerns request satisfaction; local candidates also satisfy the pixel-locality result above.

\section{Evaluation protocols}
\label{app:protocols}
\paragraph{Models and roles.} FLUX.2 [klein] 4B uses bfloat16, $512^2$, four steps, guidance one, 512 text tokens and three ordered RGB inputs on EdiVal or four on MagicBrush. Development uses seeds 42/43; transfer uses seed 44. Greedy frozen Qwen3-VL-32B supplies instruction interpretation, planning and the controller's visual checks. Evaluation uses separate models: Qwen2-VL-7B for native contextual request checks, and Qwen3-VL-8B/GLM-4.1V-9B for target-blind attribute descriptions. Qwen-Image-Edit-2511 uses bfloat16, $1024^2$, 40 steps, true classifier-free guidance six, guidance one and case--turn seeds. Controllers consume their own returned histories except in the stated shared-prefix or shared-proposal studies.

\paragraph{Supports and alignment.} Frozen float32 SAM2.1 Hiera-Small locates historical and current supports separately. Corrective images use the current mask window. Masks require score $\geq0.5$, foreground-point containment, area from four pixels to 65\% of the canvas, box sides $\geq8$ pixels, and at most 10\% outside a four-pixel box margin. Spatial restriction uses the mask union with both directional coverages $\geq0.95$ for every plausible instance. EdiVal uses a $512^2$ grid. For square MagicBrush input side $s\in\{500,1024\}$, Klein conditions on $R_d(I)$, $d=16\lfloor s/16\rfloor$, using full-extent Lanczos resizing without cropping. Composition uses aligned input $B=R_{512}(R_d(I))$; original images define semantic referents. Qwen resizes square inputs without cropping; unsupported nonsquare alignment retains the proposal.

\paragraph{Native EdiVal evaluation.} Evaluation covers 572 sequences and 1,716 positions. Nineteen content-policy holds and nineteen dependent positions reuse the preceding image: five/fourteen/nineteen holds by turn. The content-policy requests introduce no executable goals and retain their positions in the conversations and evaluation denominators. Released checks use GroundingDINO thresholds 0.3 for count and 0.35 for deletion, original overlap rules, and paired-image Qwen2-VL-7B for color/material. Cumulative success $\prod_{j\leq t}f_{i,j}$ uses issue-time request indicators $f_{i,j}$, preserving earlier failures after later recovery.

Original content consistency excludes cumulatively edited objects from unchanged-object pools, masks backgrounds from supplied-object detections, and spatially masks DINOv3 ViT-B/16 features. Means use available object/background components, first within each group and then across groups. VQ is HPSv3 reward; $\Delta_t=|N^{-1}\sum_iH(I_{i,t})-N^{-1}\sum_iH(I_{i,0})|$. Overall $O=\sqrt{\overline{\mathrm{IF}}\,\overline{\mathrm{CC}}}$ uses three-turn means. Paired 10,000-draw bootstraps cluster 570 source images, preserving trajectories, applicability and seed pairing.

\paragraph{Continuing state.} The qualified panel contains 583 earlier-request checks at 477 positions; 377 cover all nonretired requirements. Strict joint success requires complete current and earlier satisfaction. A separate 311-source panel, disjoint from development in identifiers and image content, contains 149 requests at 125 positions, eighty strict. Checks use the original introducing inputs and returned histories. With issue-time bit $a=r_g(I_j)$ and later bit $b=r_g(I_t)$, retention is $\sum ab/\sum a$ and recovery $\sum(1-a)b/\sum(1-a)$. Each controller's introducing outputs set its conditional denominators. Source bootstraps preserve all opportunities and repeated turns.

\paragraph{Controllers.} Raw uses the latest image/request; Full prompt adds raw instructions; All history also supplies past images (current first, then chronological). ReSpec retains its planning/references. Persistent goals represent live properties; Core adds them to ReSpec without spatial or local execution (shared-proposal Core is $A$). Static protection dilates SAM unions by four pixels and feathers by two pixels. Adapted SEAM retains state memory, resource-aware retrieval, graph expansion and applicable-fact filtering. Internal variants share interpretations; external controls plan their own proposals. Common token limits retain complete goals/reference order. Complete benchmarks use the same TRACE executor with goal-dependent local branches. The color--material scope for acquisition and refinement occurs at two EdiVal positions per editor/seed setting and none on MagicBrush, whose requests span broader edit types. Controlled color--material studies evaluate these components directly.

\paragraph{RISE-Agent (adapted).} We use Qwen3-VL-32B for planning/verification and Klein 4B in place of GPT-5.1/Gemini-3-Flash and Klein 9B. Each material turn receives the actual shared $H$ and current request only. A uniform counterfactual task label, fixed before editing, exempts intended unusual materials from realism penalties while retaining correctness, artifact and consistency checks. We retain tools, one replan, two refinements and at most four candidates. The native seed rule combines global seed 42 with cohort, input-replicate seed, case, turn and attempt. Native scores select the best fully evaluated candidate, with ties favoring earlier attempts; when none is available, the input $H$ is retained.

\paragraph{Retrieval.} BM25 uses lowercase words, positive inverse document frequency, $k_1=1.2$, $b=0.75$ and top-one selection. Its Full prompt template receives selected/current requests and the current image. CLIP ViT-B/32 ranks actual history images by current-request cosine similarity, retaining the raw instruction prefix, current image first and at most one distinct reference. Ties favor recency. With one earlier request, BM25 equals Full prompt; CLIP yields Full prompt or All history.

Spatial restriction is disabled in the controlled appearance studies and compared with unrestricted editing in the eight-turn study.

\section{Complete benchmark and spatial results}
\label{app:legacy-results}
\begin{table*}[!t]
\centering\footnotesize
\setlength{\tabcolsep}{3.5pt}
\begin{tabular}{lrrrrr}
\toprule
Controller & IF$_{1/2/3}$ & mIF$_{1/2/3}$ & CC$_{1/2/3}$ & VQ$_{1/2/3}$ & $\Delta_{1/2/3}$ \\
\midrule
\multicolumn{6}{l}{\textit{Klein, seed 42}} \\
Raw editor & 66.96 / 42.83 / 26.92 & 66.96 / 65.73 / 65.73 & 95.75 / 92.28 / 89.49 & \textbf{6.53} / 6.67 / 6.49 & 2.14 / 2.28 / 2.10 \\
Full prompt & 66.96 / 41.26 / 22.90 & 66.96 / 63.64 / 58.74 & 95.75 / 92.16 / 89.46 & \textbf{6.53} / 6.67 / 6.40 & 2.14 / 2.28 / 2.01 \\
All history & 66.96 / 39.69 / 22.38 & 66.96 / 61.19 / 57.52 & 95.75 / \textbf{93.55} / \textbf{92.19} & \textbf{6.53} / 6.46 / 6.13 & 2.14 / \textbf{2.08} / \textbf{1.75} \\
ReSpec & 66.96 / 47.38 / 29.72 & 66.96 / 71.68 / 68.01 & 95.75 / 91.83 / 88.64 & \textbf{6.53} / 6.76 / 6.56 & 2.14 / 2.37 / 2.18 \\
Adapted SEAM & \textbf{67.31} / 41.96 / 27.27 & \textbf{67.31} / 64.51 / 66.78 & 95.44 / 92.03 / 89.15 & \textbf{6.53} / 6.65 / 6.47 & 2.14 / 2.27 / 2.08 \\
BM25 history & 66.96 / 41.26 / 23.95 & 66.96 / 63.64 / 60.49 & 95.75 / 92.16 / 89.57 & \textbf{6.53} / 6.67 / 6.50 & 2.14 / 2.28 / 2.11 \\
CLIP retrieval & 66.96 / 40.73 / 22.90 & 66.96 / 62.76 / 59.62 & 95.75 / 92.79 / 90.88 & \textbf{6.53} / 6.58 / 6.27 & 2.14 / 2.20 / 1.88 \\
Persistent goals & 66.96 / 44.41 / 29.20 & 66.96 / 67.48 / 66.78 & 95.75 / 91.55 / 88.23 & \textbf{6.53} / 6.62 / 6.54 & 2.14 / 2.23 / 2.15 \\
Core & 66.96 / \textbf{48.25} / 30.77 & 66.96 / \textbf{72.90} / 68.71 & 95.75 / 91.60 / 88.37 & \textbf{6.53} / \textbf{6.78} / \textbf{6.60} & 2.14 / 2.40 / 2.22 \\
TRACE & 66.96 / 48.08 / \textbf{31.12} & 66.96 / 72.73 / \textbf{69.41} & \textbf{95.81} / 91.64 / 88.44 & 6.51 / 6.77 / \textbf{6.60} & \textbf{2.12} / 2.39 / 2.22 \\
\midrule
\multicolumn{6}{l}{\textit{Klein, seed 43}} \\
TRACE & 67.31 / 46.33 / 30.94 & 67.31 / 70.80 / 68.53 & 95.77 / 91.69 / 88.61 & 6.64 / 6.94 / 6.80 & 2.26 / 2.55 / 2.41 \\
\midrule
\multicolumn{6}{l}{\textit{Qwen-Image-Edit-2511}} \\
Raw editor & \textbf{77.45} / 56.82 / 40.73 & \textbf{77.45} / 75.52 / 73.95 & 93.62 / 89.65 / 86.71 & \textbf{6.47} / 6.37 / \textbf{6.30} & \textbf{2.08} / \textbf{1.98} / 1.91 \\
ReSpec & \textbf{77.45} / 57.34 / 42.31 & \textbf{77.45} / 76.22 / 73.60 & 93.62 / \textbf{89.82} / \textbf{86.89} & \textbf{6.47} / 6.46 / 6.24 & \textbf{2.08} / 2.08 / \textbf{1.85} \\
Source-qualified core & \textbf{77.45} / 57.69 / 42.83 & \textbf{77.45} / 76.40 / \textbf{75.70} & 93.62 / 89.77 / 86.68 & \textbf{6.47} / \textbf{6.47} / 6.26 & \textbf{2.08} / 2.08 / 1.87 \\
TRACE & \textbf{77.45} / \textbf{57.87} / \textbf{43.01} & \textbf{77.45} / \textbf{76.57} / \textbf{75.70} & \textbf{93.67} / 89.80 / 86.72 & \textbf{6.47} / 6.46 / 6.26 & \textbf{2.08} / 2.07 / 1.87 \\
\multicolumn{6}{l}{Overall $O$ (\%): Raw 72.45, ReSpec 72.94, Core 73.08, TRACE \textbf{73.17}.} \\
\bottomrule
\end{tabular}
\caption{Complete native EdiVal means on all 572 sequences and common holds. Triples give turns one--three; IF/mIF/CC are percentages, VQ/$\Delta$ use HPSv3 reward units. Bold marks best values and ties within editor/seed (IF, mIF, CC, VQ: higher; $\Delta$: lower). Definitions/applicability follow Sec.~\ref{app:protocols}.}
\label{tab:all-per-turn}
\end{table*}

\begin{table}[t]
\centering\small
\setlength{\tabcolsep}{3pt}
\begin{tabular*}{\columnwidth}{@{\extracolsep{\fill}}lrrrr}
\toprule
&\multicolumn{2}{c}{Full}&\multicolumn{2}{c}{Source-disjoint}\\
\cmidrule(lr){2-3}\cmidrule(lr){4-5}
Controller & Joint & Strict & Joint & Strict\\
Denominator &477&377&125&80\\
\midrule
\multicolumn{5}{l}{\textit{Klein, seed 42}}\\
ReSpec &199&156&\textbf{64}&\textbf{39}\\
Core &\textbf{211}&\textbf{169}&63&\textbf{39}\\
TRACE &208&166&60&36\\
\midrule
\multicolumn{5}{l}{\textit{Qwen-Image-Edit-2511}}\\
ReSpec &263&204&83&55\\
Core &264&\textbf{210}&\textbf{84}&\textbf{57}\\
TRACE &\textbf{265}&\textbf{210}&\textbf{84}&\textbf{57}\\
\bottomrule
\end{tabular*}
\caption{Continuing-state success on actual returned histories. Joint requires the current and all qualified earlier requests; Strict requires complete coverage of nonretired earlier requests. The fixed source-disjoint panel excludes development sources. All controllers follow their own returned histories.}
\label{tab:continuing-state}
\end{table}

\paragraph{Continuing-state checks.} These checks credit recovered earlier requirements, complementing cumulative IF and targeted recovery studies. On the source-disjoint panel, TRACE ties Core with Qwen and trails ReSpec and Core with Klein (Table~\ref{tab:continuing-state}).

\paragraph{MagicBrush.} The six context baselines coincide when independent edits reset to supplied inputs; own-history sessions consume returned outputs. All 1,053 edits and 535 final images retain original DINO ViT-S/16 and CLIP ViT-B/32 metrics (Table~3 of the main paper). Bootstrap draws cluster sessions.

\paragraph{Spatial selection.} With proposals, masks and removal/overlap checks fixed, complete-request selection preserves current/joint counts while improving consistency. Unconditional restriction may erase edits; current-only checking may admit earlier regressions.

\paragraph{Coverage sensitivity.} Across 916 fixed positions, stricter directional coverage admits fewer restrictions with unchanged current/continuing answers; exact equality admits none. All complete benchmarks use 0.95.

\section{Attribute-specific recovery}
\label{app:material-recovery}
\paragraph{Conversations and readouts.} The 56 development requests include 54 color edits and two replacements partly superseded by the material turn, retaining five holds. \mbox{Earlier/Joint} uses 54 requests per seed; Current and appearance retain all 56 outputs. Material targets are metal, or wood when already metal. Native E/C/J denotes Earlier/Current/Joint; blind C/M/P denotes Color/Material/Pair. Entries $t^{u}$ report positive/unknown counts; negatives number $n-t-u$.

Qwen3-VL-8B and GLM-4.1V-9B observe neutral-category color/material, with targets bound afterward. Ambiguous, unavailable, multicolor and invalid labels remain $U$; gray/grey are normalized. Pair is $T$ for two positives, $F$ for any failure, otherwise $U$. GLM uses its official RGB processor, bfloat16, greedy decoding and 8,192 tokens. 

\paragraph{Interpreting the readouts.} Native checks assess complete requests against introducing inputs. Blind readers describe single-image body color/material from neutral categories; targets are matched afterward. An image may pass contextual editing yet receive an unknown body-color label, or match appearance while failing another clause, such as identity. Table~\ref{tab:reader-target-check} reports intended-target agreement on released MagicBrush images.

\begin{table}[!t]
\centering\small
\setlength{\tabcolsep}{2pt}
\begin{tabular}{@{}p{67pt}rrr@{}}
\toprule
Method & Native & 8B & GLM\\
& E/C/J & C/M/P & C/M/P\\
\midrule
Raw editor & 64/\textbf{110}/62 & $18^{21}/86^{2}/13^{20}$ & $11^{14}/85^{4}/9^{12}$\\
Full prompt & 82/107/78 & $30^{17}/78^{1}/19^{16}$ & $16^{19}/78^{6}/10^{15}$\\
All history & 89/106/83 & $53^{16}/74^{3}/28^{14}$ & $44^{18}/71^{9}/22^{15}$\\
ReSpec & 84/106/79 & $27^{27}/76^{2}/17^{23}$ & $18^{29}/73^{10}/8^{28}$\\
Adapted SEAM & 64/\textbf{110}/62 & $18^{20}/86^{2}/13^{19}$ & $11^{14}/85^{4}/9^{12}$\\
RISE-Agent (adapted) & 69/108/66 & $26^{29}/\mathbf{88}^{2}/16^{29}$ & $20^{22}/\mathbf{88}^{8}/17^{19}$\\
BM25 history & 82/107/78 & $30^{17}/78^{1}/19^{16}$ & $16^{19}/78^{6}/10^{15}$\\
CLIP retrieval & 88/106/82 & $42^{17}/75^{1}/25^{15}$ & $29^{20}/72^{8}/16^{15}$\\
Core & 88/\textbf{110}/87 & $39^{24}/77^{3}/22^{24}$ & $31^{24}/79^{8}/18^{23}$\\
History pixels & 98/92/79 & $62^{24}/59^{3}/21^{22}$ & $57^{23}/56^{10}/18^{22}$\\
History chroma & 95/109/93 & $53^{25}/69^{3}/26^{25}$ & $47^{21}/73^{9}/29^{21}$\\
Chroma-only & 95/109/93 & $59^{22}/70^{3}/35^{22}$ & $53^{21}/75^{8}/36^{20}$\\
No evidence draw & 94/\textbf{110}/93 & $53^{23}/69^{3}/28^{23}$ & $47^{22}/73^{8}/28^{21}$\\
No color check & 95/\textbf{110}/94 & $58^{25}/69^{3}/31^{25}$ & $51^{21}/72^{9}/32^{21}$\\
Current-only & 93/\textbf{110}/92 & $52^{25}/71^{3}/30^{25}$ & $46^{21}/75^{8}/31^{20}$\\
Color, affine & 95/\textbf{110}/94 & $60^{22}/68^{3}/34^{22}$ & $54^{21}/73^{8}/35^{20}$\\
Color + half-step & 95/\textbf{110}/94 & $60^{22}/72^{3}/38^{22}$ & $52^{23}/74^{8}/34^{22}$\\
Two half-steps & 94/\textbf{110}/93 & $60^{22}/72^{3}/38^{22}$ & $50^{24}/74^{8}/32^{23}$\\
Material veto & 96/\textbf{110}/95 & $59^{22}/68^{3}/33^{22}$ & $54^{20}/75^{7}/36^{19}$\\
Acquired copy & 96/\textbf{110}/95 & $64^{19}/70^{2}/38^{20}$ & $59^{18}/75^{9}/40^{20}$\\
Raw expressions & 98/\textbf{110}/97 & $63^{20}/68^{2}/35^{20}$ & $58^{18}/79^{7}/42^{18}$\\
No carrier check & \textbf{99}/109/97 & $67^{21}/68^{2}/35^{20}$ & $63^{21}/73^{9}/40^{21}$\\
Carrier, affine & 98/\textbf{110}/97 & $65^{19}/69^{2}/37^{20}$ & $60^{18}/77^{8}/42^{19}$\\
Uncalibrated taper & 98/109/96 & $65^{19}/72^{2}/40^{20}$ & $57^{20}/78^{8}/41^{21}$\\
Color + calibration & 95/109/93 & $60^{22}/71^{3}/37^{22}$ & $53^{21}/75^{8}/37^{20}$\\
Single proposal & 98/109/96 & $65^{19}/73^{2}/\mathbf{41}^{20}$ & $59^{18}/79^{8}/44^{19}$\\
TRACE & 98/109/96 & $65^{19}/73^{2}/\mathbf{41}^{20}$ & $59^{18}/80^{8}/\mathbf{45}^{19}$\\
Whole carrier, aff. & \textbf{99}/\textbf{110}/\textbf{98} & $69^{19}/70^{2}/37^{20}$ & $64^{19}/74^{8}/42^{19}$\\
Whole all, aff. & \textbf{99}/109/97 & $\mathbf{70}^{19}/69^{2}/37^{20}$ & $\mathbf{66}^{18}/73^{8}/43^{18}$\\
Whole carrier, cal. & \textbf{99}/109/97 & $69^{19}/73^{2}/40^{20}$ & $63^{19}/76^{8}/44^{19}$\\
Whole all, calib. & \textbf{99}/108/96 & $\mathbf{70}^{19}/72^{2}/40^{20}$ & $65^{18}/75^{8}/\mathbf{45}^{18}$\\
Checked global & 96/109/94 & $62^{19}/74^{2}/40^{20}$ & $56^{18}/79^{7}/43^{17}$\\
\bottomrule
\end{tabular}
\caption{All 112 development outputs from 56 sources and two seeds. Native E/C/J uses 108/112/108 positions; blind C/M/P is $T^{U}$ (remaining $F$). Externals share T1; operators share $A$. Component controls permit at most one corrective image; carrier/all global controls use 37/45 draws. Bold marks maxima and ties.}
\label{tab:blind8b}
\label{tab:local-modules-full}
\end{table}

\begin{table}[!t]
\centering\small
\setlength{\tabcolsep}{2pt}
\begin{tabular}{lrrr}
\toprule
Method & Native & 8B & GLM\\
& E/C/J & C/M/P & C/M/P\\
\midrule
\multicolumn{4}{l}{\textit{Klein, seed 44: all 56}}\\
Core & 40/\textbf{52}/37 & $18^{13}/\mathbf{42}^{2}/12^{14}$ & $15^{11}/\mathbf{39}^{2}/8^{11}$\\
History pixels & \textbf{47}/42/34 & $\mathbf{31}^{12}/29^{2}/10^{12}$ & $\mathbf{30}^{11}/29^{2}/9^{10}$\\
History chroma & 46/\textbf{52}/\textbf{43} & $26^{12}/38^{2}/15^{13}$ & $26^{10}/\mathbf{39}^{2}/\mathbf{16}^{10}$\\
Color executor & 46/\textbf{52}/\textbf{43} & $29^{11}/37^{2}/\mathbf{18}^{12}$ & $26^{10}/37^{2}/15^{10}$\\
Color + half-step & 46/\textbf{52}/\textbf{43} & $27^{13}/38^{2}/17^{14}$ & $24^{10}/36^{2}/13^{10}$\\
Affine carrier & 46/\textbf{52}/\textbf{43} & $30^{11}/36^{2}/\mathbf{18}^{12}$ & $26^{11}/36^{3}/15^{11}$\\
Local-only & 46/\textbf{52}/\textbf{43} & $30^{11}/35^{2}/17^{12}$ & $26^{11}/37^{3}/\mathbf{16}^{11}$\\
\midrule
\multicolumn{4}{l}{\textit{Qwen-Image-Edit-2511: all 56}}\\
Core & 46/\textbf{54}/44 & $28^{9}/\mathbf{46}^{0}/23^{8}$ & $29^{4}/\mathbf{50}^{1}/28^{3}$\\
History pixels & 46/53/43 & $32^{9}/43^{0}/23^{8}$ & $34^{3}/47^{0}/28^{3}$\\
History chroma & 46/\textbf{54}/44 & $31^{9}/\mathbf{46}^{0}/25^{8}$ & $32^{3}/\mathbf{50}^{0}/30^{3}$\\
Color executor & \textbf{47}/\textbf{54}/\textbf{45} & $33^{9}/\mathbf{46}^{0}/\mathbf{27}^{8}$ & $34^{4}/49^{1}/32^{3}$\\
Color + half-step & \textbf{47}/\textbf{54}/\textbf{45} & $32^{9}/\mathbf{46}^{0}/26^{8}$ & $34^{4}/49^{1}/32^{3}$\\
Whole render & \textbf{47}/53/44 & $\mathbf{36}^{8}/44^{0}/\mathbf{27}^{7}$ & $\mathbf{37}^{3}/47^{0}/31^{3}$\\
Corrective copy & \textbf{47}/53/44 & $35^{8}/44^{0}/26^{7}$ & $\mathbf{37}^{3}/47^{0}/31^{3}$\\
Affine carrier & \textbf{47}/\textbf{54}/\textbf{45} & $34^{8}/45^{0}/26^{8}$ & $35^{4}/49^{1}/\mathbf{33}^{3}$\\
Local-only & \textbf{47}/\textbf{54}/\textbf{45} & $34^{8}/45^{0}/26^{8}$ & $34^{5}/49^{1}/32^{4}$\\
\bottomrule
\end{tabular}
\caption{Local-subsystem transfer with global fallback and retry disabled. Each panel has 56 outputs; native E/J uses 54 applicable requests. Blind C/M/P reports $T^{U}$ (remaining $F$). Each editor uses its own first outputs and sampler; internal controls share $A$ and at most one corrective draw. Bold marks maxima and ties per panel.}
\label{tab:seed-editor-transfer}
\end{table}

\begin{table*}[!t]
\centering\small
\setlength{\tabcolsep}{4pt}
\begin{tabular}{lrrrrrrrrr}
\toprule
 & \multicolumn{3}{c}{Original requests} & \multicolumn{3}{c}{Qwen3-VL-8B} & \multicolumn{3}{c}{GLM-4.1V-9B} \\
\cmidrule(lr){2-4}\cmidrule(lr){5-7}\cmidrule(lr){8-10}
Policy & Earlier & Material & Both & Color & Material & Pair & Color & Material & Pair \\
\midrule
\multicolumn{10}{l}{\textit{Matched Klein sampler and complete controller: seeds 42/43, $n=112$}} \\
Observed + half-step & \textbf{95} & \textbf{110} & \textbf{94} & \textbf{60} & \textbf{72} & \textbf{38} & \textbf{52} & 74 & \textbf{34} \\
CSS + half-step & \textbf{95} & \textbf{110} & \textbf{94} & 57 & 71 & 35 & 51 & \textbf{75} & \textbf{34} \\
\midrule
\multicolumn{10}{l}{\textit{Direct color-editing baseline: all 56 sources at seed 42}} \\
Core & 46 & \textbf{54} & 45 & 18 & 37 & 9 & 14 & \textbf{39} & 8 \\
ColorCtrl & 45 & 51 & 42 & 26 & 27 & 11 & 21 & 32 & 11 \\
ColorCtrl + goal checks & 46 & \textbf{54} & 45 & 20 & \textbf{38} & 12 & 16 & \textbf{39} & 9 \\
Color executor & 48 & \textbf{54} & 47 & 27 & 33 & 14 & 24 & 36 & 15 \\
Affine carrier executor & \textbf{49} & \textbf{54} & \textbf{48} & \textbf{29} & 34 & 15 & \textbf{25} & \textbf{39} & \textbf{18} \\
Color executor + half-step & 48 & \textbf{54} & 47 & 27 & 35 & \textbf{16} & 23 & 37 & 15 \\
\bottomrule
\end{tabular}
\caption{Matched target-center controls (112 outputs) and released ColorCtrl comparison (all 56 seed-42 outputs). Entries are positive counts; native Earlier/Joint use 108 or 54 applicable requests. CSS and observed centers share one half-step. ColorCtrl uses released FLUX.1-dev, $768^2$, 50 inversion/50 denoising steps; Klein uses $512^2$ and four steps. Goal checks retain $A$ on failure. Bold marks best counts and ties within each panel.}
\label{tab:matched-colors}
\end{table*}

\begin{table}[!t]
\centering\small
\setlength{\tabcolsep}{1.5pt}
\begin{tabular}{@{}p{67pt}rrr@{}}
\toprule
Method & Native & 8B & GLM\\
& E/C/J & C/M/P & C/M/P\\
\midrule
Raw editor & 59/111/52 & $13^{20}/84^{3}/3^{18}$ & $9^{16}/92^{7}/3^{11}$\\
Full prompt & 80/109/73 & $39^{16}/73^{3}/16^{12}$ & $35^{13}/79^{13}/14^{14}$\\
All history & 84/102/68 & $46^{18}/68^{2}/17^{13}$ & $40^{18}/67^{19}/12^{18}$\\
ReSpec & 68/112/63 & $24^{18}/73^{5}/13^{14}$ & $19^{14}/90^{10}/10^{13}$\\
BM25 history & 80/109/73 & $39^{16}/73^{3}/16^{12}$ & $35^{13}/79^{13}/14^{14}$\\
CLIP retrieval & 83/107/73 & $41^{17}/73^{2}/16^{13}$ & $38^{12}/75^{15}/13^{14}$\\
RISE-Agent (adapted) & 66/\textbf{115}/62 & $13^{18}/\mathbf{101}^{1}/8^{16}$ & $7^{11}/\mathbf{103}^{8}/4^{8}$\\
Core & 71/113/67 & $21^{24}/80^{4}/8^{20}$ & $15^{14}/92^{9}/8^{11}$\\
Color, affine & 82/112/77 & $42^{22}/73^{3}/20^{18}$ & $36^{14}/85^{10}/22^{12}$\\
Color + calib. & 82/112/77 & $42^{22}/74^{3}/21^{18}$ & $36^{14}/86^{10}/23^{12}$\\
Carrier, affine & 81/113/77 & $44^{22}/73^{3}/21^{18}$ & $37^{14}/86^{9}/23^{12}$\\
Local-only & 81/112/76 & $44^{22}/74^{3}/22^{18}$ & $37^{14}/87^{9}/24^{12}$\\
Single proposal & 83/112/78 & $46^{21}/74^{3}/24^{17}$ & $39^{14}/87^{9}/26^{12}$\\
TRACE & 83/113/79 & $46^{21}/75^{3}/25^{17}$ & $39^{14}/88^{9}/\mathbf{27}^{12}$\\
Whole all, aff. & \textbf{104}/94/\textbf{80} & $\mathbf{73}^{18}/57^{3}/26^{15}$ & $\mathbf{66}^{14}/62^{11}/24^{15}$\\
Whole all, calib. & \textbf{104}/93/79 & $\mathbf{73}^{18}/58^{3}/\mathbf{27}^{15}$ & $\mathbf{66}^{14}/63^{11}/25^{15}$\\
Checked global & 83/112/78 & $45^{22}/75^{4}/25^{18}$ & $38^{14}/88^{9}/26^{12}$\\
\bottomrule
\end{tabular}
\caption{All 120 COCO positions: native E/C/J and blind C/M/P $T^{U}$ (remaining $F$). Externals share T1; internal controls share $A$ with at most one corrective image. TRACE adds the qualified retry. Checked-global selection is specified in Sec.~\ref{app:local-modules}. Bold marks maxima and ties.}
\label{tab:coco120-full}
\label{tab:external-transfer-full}
\end{table}

\begin{table}[!t]
\centering\small
\setlength{\tabcolsep}{2pt}
\begin{tabular}{lrrr}
\toprule
Method & Native & 8B & GLM\\
& E/C/J & C/M/P & C/M/P\\
\midrule
Swatch, local & 98/\textbf{109}/96 & $63^{19}/\mathbf{71}^{1}/37^{19}$ & $57^{17}/79^{8}/42^{18}$\\
History, local & 97/\textbf{109}/95 & $61^{20}/70^{2}/36^{20}$ & $56^{18}/\mathbf{80}^{8}/43^{19}$\\
Swatch, global & \textbf{100}/108/\textbf{97} & $\mathbf{70}^{19}/70^{1}/\mathbf{38}^{19}$ & $\mathbf{66}^{17}/76^{7}/\mathbf{47}^{16}$\\
History, global & \textbf{100}/105/94 & $69^{20}/69^{2}/37^{20}$ & $63^{21}/72^{7}/43^{17}$\\
\bottomrule
\end{tabular}
\caption{Corrective-reference ablations on 112 development positions, with notation and denominators from Table~\ref{tab:blind8b}. Both arms add a reference image and identical color-only cue to the Local-only rendering prompt; reference content is the sole difference. Execution and calibration are fixed. The local arms modify 37 routes; global rows return all 45 draws, including eight unchanged originals. Bold marks within-table maxima and ties.}
\label{tab:corrective-reference}
\end{table}

\begin{table}[!t]
\centering\small
\setlength{\tabcolsep}{4pt}
\begin{tabular}{lrrr}
\toprule
Scope & $N$ & 8B T/F/U & GLM T/F/U\\
\midrule
Complete &62&44/16/2&39/18/5\\
Prespecified text scope &51&38/11/2&33/14/4\\
\bottomrule
\end{tabular}
\caption{Intended-target agreement on released MagicBrush outputs, with requested attributes as reference labels. The 52 color/10 material requests yield 8B T/F/U counts of 35/15/2 and 9/1/0; GLM yields 32/15/5 and 7/3/0. Both readers cover all 62 outputs; seven unknown answers concern multicolored objects. Readers receive the image and neutral category.}
\label{tab:reader-target-check}
\end{table}

\begin{table*}[!t]
\centering\small
\setlength{\tabcolsep}{4pt}
\begin{tabular}{lrrrr}
\toprule
Policy & Object feature & Object pixel & Background feature & Background pixel\\
\midrule
Color executor (affine) & 91.95 & 93.28 & \textbf{97.43} & 96.17\\
Affine carrier executor & \textbf{91.96} & \textbf{93.31} & \textbf{97.43} & 96.17\\
Local-only & 91.95 & 93.30 & \textbf{97.43} & 96.17\\
Single proposal & 91.87 & 93.24 & 97.40 & 96.10\\
TRACE & 91.86 & 93.24 & 97.40 & 96.10\\
Whole acquired carrier (affine) & 91.56 & 93.24 & 96.57 & 95.93\\
Whole corrective render (affine) & 90.75 & 92.89 & 96.46 & 95.18\\
Whole acquired carrier (calibrated) & 91.55 & 93.23 & 96.57 & 95.93\\
Whole corrective render (calibrated) & 90.74 & 92.89 & 96.46 & 95.18\\
Checked global & 91.79 & 93.16 & 96.55 & 96.00\\
RISE-Agent (adapted) & 90.07 & 92.67 & 97.05 & \textbf{96.59}\\
\bottomrule
\end{tabular}
\caption{COCO120 consistency to shared T1 ($H$; $\times100$, higher). Executor controls use a one-draw cap; TRACE includes retry. Objects: 103 sources (17 structural NAs); background features: 120; pixels: 119/120. Means use available values. Detector/DINOv3 pools use source annotations and replacement classes independently of editing masks. Bold marks displayed maxima and ties.}
\label{tab:context-cc}
\end{table*}

\begin{table}[!t]
\centering\small
\setlength{\tabcolsep}{2pt}
\begin{tabular}{@{}p{67pt}rrr@{}}
\toprule
Method & Native & 8B & GLM\\
& E/C/J & C/M/P & C/M/P\\
\midrule
\multicolumn{4}{l}{\textit{Affine goal-state/external controls}}\\
Raw (recent) & 80/94/71 & $49^{8}/\mathbf{34}^{7}/7^{6}$ & $42^{14}/37^{5}/6^{7}$\\
Full prompt & 83/93/71 & $59^{10}/24^{6}/9^{7}$ & $56^{19}/25^{13}/6^{14}$\\
Prompt rewrite & \textbf{94}/90/73 & $\mathbf{93}^{2}/24^{3}/16^{3}$ & $82^{13}/22^{14}/12^{14}$\\
ReSpec & 84/97/70 & $62^{6}/28^{6}/7^{7}$ & $55^{10}/32^{5}/6^{6}$\\
BM25 history & 79/99/69 & $51^{6}/33^{7}/10^{6}$ & $42^{18}/39^{10}/8^{11}$\\
CLIP retrieval & 83/93/72 & $64^{3}/22^{3}/6^{1}$ & $60^{11}/26^{10}/8^{11}$\\
Accumulated & 78/94/68 & $76^{4}/31^{5}/17^{5}$ & $66^{17}/36^{5}/18^{8}$\\
Color, active & 86/\textbf{100}/\textbf{78} & $77^{6}/31^{4}/\mathbf{18}^{6}$ & $70^{14}/39^{5}/21^{6}$\\
\midrule
\multicolumn{4}{l}{\textit{Calibrated local-executor controls}}\\
Color, off & 86/\textbf{100}/\textbf{78} & $76^{6}/31^{4}/17^{6}$ & $71^{14}/\mathbf{40}^{5}/23^{6}$\\
Carrier, off & 85/99/75 & $77^{8}/29^{5}/15^{7}$ & $73^{14}/\mathbf{40}^{5}/\mathbf{24}^{6}$\\
Global, off & 92/94/76 & $\mathbf{93}^{6}/21^{5}/15^{7}$ & $89^{13}/28^{7}/19^{9}$\\
Color, on & 85/\textbf{100}/77 & $77^{5}/29^{5}/16^{6}$ & $73^{14}/37^{6}/22^{7}$\\
Carrier, on & 84/\textbf{100}/75 & $76^{8}/28^{5}/14^{7}$ & $73^{16}/38^{5}/21^{8}$\\
Global, on & 92/96/\textbf{78} & $91^{6}/22^{5}/14^{7}$ & $\mathbf{90}^{13}/25^{7}/18^{8}$\\
\bottomrule
\end{tabular}
\caption{Sixteen eight-turn histories: native E/C/J uses 109/112/109 positions; blind C/M/P is $T^{U}$ over 112. Calibrated controls disable fallback/retry. On/off toggles spatial scope; Global returns its draw, otherwise local output. Bold marks maxima/ties.}
\label{tab:goal-lifecycle-full}
\label{tab:lifecycle-calibrated}
\end{table}

\begin{table}[!t]
\centering\small
\setlength{\tabcolsep}{2pt}
\begin{tabular}{lrrrr}
\toprule
Policy & Object F & Object P & Bg. F & Bg. P\\
\midrule
Full prompt & 0.8641 & 0.8860 & 0.9827 & 0.9437\\
ReSpec & 0.8615 & 0.8924 & 0.9829 & 0.9514\\
Color, calibrated & \textbf{0.8851} & \textbf{0.9063} & \textbf{0.9877} & \textbf{0.9646}\\
Whole corrective & 0.8409 & 0.8851 & 0.9823 & 0.9406\\
TRACE & 0.8799 & 0.9042 & 0.9875 & 0.9629\\
\bottomrule
\end{tabular}
\caption{Independent32 context against T1 across both later turns; F/P denotes feature/pixel consistency (higher). Object $n=56$ positions/28 sources; background $n=64$. Bold marks maxima.}
\label{tab:independent32-context}
\end{table}

\begin{table}[t]
\centering\small
\setlength{\tabcolsep}{2pt}
\renewcommand{\arraystretch}{0.96}
\begin{tabular}{lrrr}
\toprule
Method & Native & 8B & GLM\\
& E/C/J & C/M/P & C/M/P\\
\midrule
\multicolumn{4}{l}{\textit{All later outputs ($n=64$)}}\\
Full prompt & $42/58/36$ & $37^{4}/34^{3}/16^{5}$ & $37^{5}/35^{6}/16^{7}$\\
ReSpec & $39/\mathbf{61}/37$ & $33^{5}/37^{1}/17^{5}$ & $33^{3}/\mathbf{43}^{1}/21^{2}$\\
Color, calibrated & $38/\mathbf{61}/36$ & $37^{4}/37^{3}/16^{6}$ & $37^{3}/35^{8}/18^{4}$\\
Whole corrective & $\mathbf{54}/56/\mathbf{46}$ & $\mathbf{43}^{4}/\mathbf{38}^{3}/\mathbf{22}^{5}$ & $\mathbf{43}^{2}/36^{7}/\mathbf{22}^{4}$\\
TRACE & $39/\mathbf{61}/37$ & $37^{4}/37^{3}/16^{6}$ & $37^{4}/34^{9}/18^{5}$\\
\midrule
\multicolumn{4}{l}{\textit{Material turn ($n=32$)}}\\
Full prompt & $21/26/15$ & $9^{3}/20^{2}/5^{3}$ & $8^{4}/23^{2}/5^{3}$\\
ReSpec & $19/\mathbf{30}/18$ & $5^{4}/\mathbf{21}^{0}/4^{3}$ & $4^{3}/\mathbf{24}^{0}/4^{1}$\\
Color, calibrated & $19/29/17$ & $8^{3}/\mathbf{21}^{2}/3^{4}$ & $7^{2}/19^{5}/2^{2}$\\
Whole corrective & $\mathbf{27}/26/\mathbf{21}$ & $\mathbf{21}^{2}/18^{2}/\mathbf{10}^{3}$ & $\mathbf{20}^{2}/16^{6}/\mathbf{8}^{4}$\\
TRACE & $19/29/17$ & $8^{3}/\mathbf{21}^{2}/3^{4}$ & $7^{3}/18^{6}/2^{3}$\\
\midrule
\multicolumn{4}{l}{\textit{Terminal color revision ($n=32$)}}\\
Full prompt & $21/\mathbf{32}/21$ & $28^{1}/14^{1}/11^{2}$ & $29^{1}/12^{4}/11^{4}$\\
ReSpec & $20/31/19$ & $28^{1}/16^{1}/\mathbf{13}^{2}$ & $29^{0}/19^{1}/\mathbf{17}^{1}$\\
Color, calibrated & $19/\mathbf{32}/19$ & $\mathbf{29}^{1}/16^{1}/\mathbf{13}^{2}$ & $\mathbf{30}^{1}/16^{3}/16^{2}$\\
Whole corrective & $\mathbf{27}/30/\mathbf{25}$ & $22^{2}/\mathbf{20}^{1}/12^{2}$ & $23^{0}/\mathbf{20}^{1}/14^{0}$\\
TRACE & $20/\mathbf{32}/20$ & $\mathbf{29}^{1}/16^{1}/\mathbf{13}^{2}$ & $\mathbf{30}^{1}/16^{3}/16^{2}$\\
\bottomrule
\end{tabular}
\caption{Independent three-turn conversations with balanced colors and sixteen wood/metal targets each. Native E/C/J is complete Earlier/Current/Joint (Earlier: color at T2, material at T3); blind C/M/P is $T^{U}$, remaining $F$. Only T1 is shared; later turns follow own histories. Bold marks within-panel maxima/ties.}
\label{tab:independent32-full}
\end{table}
\subsection{Execution, carrier and calibration controls}
\label{app:local-modules}
\emph{History pixels} restores the historical region; \emph{History chroma} translates current chroma using history without generation. \emph{Chroma-only} fixes the family and disables analytic gray. \emph{No evidence draw} omits generation; \emph{No color check} omits color and complete-request checks; \emph{Current-only} checks the current request. \emph{Material veto} requires positive neutral material; \emph{Acquired copy} omits color execution on $R$; \emph{Raw expressions} repeats historical wording; \emph{No carrier check} omits final acquisition checks. Evidence, admission, current-only and half-step controls use color-only execution; Chroma-only and carrier controls retain acquisition.

Operator and reference controls share $A$ with at most one corrective image; global fallback and retry are disabled. The factorial varies acquisition and calibration on this common input. \emph{Local-only} is its complete configuration; \emph{Single proposal} also applies the checked global fallback. TRACE additionally permits the second material proposal.

\paragraph{Global controls.} Whole rendering returns the first proposal's $R$ globally. Checked global requires neutral color and all complete active requests to pass; otherwise it returns $B$ ($A$ or $P$). Both return the local result when no corrective image is generated. These controls isolate editing scope within one proposal (Tables~\ref{tab:coco120-full} and~\ref{tab:context-cc}).

\begin{table}[!t]
\centering\small
\setlength{\tabcolsep}{3pt}
\begin{tabular}{@{}lrrrr@{}}
\toprule
Executor & Queries $\downarrow$ & Masks $\downarrow$ & Draws $\downarrow$ & Pair $\uparrow$\\
\midrule
Color, affine &\textbf{353}&\textbf{74}&\textbf{19}&34/35\\
Color, calibrated &390&\textbf{74}&\textbf{19}&37/37\\
Carrier, affine &772&117&45&37/42\\
Local-only full &809&117&45&\textbf{41}/44\\
Single proposal &834&117&45&\textbf{41}/44\\
TRACE &1138&117&79&\textbf{41/45}\\
\bottomrule
\end{tabular}
\caption{Development repair cost over 112 inputs, excluding ordinary editing, spatial selection and independent evaluation. Queries count logical attribute/request and grounding calls; Masks count mask-request contexts; Draws count corrective images. Pair is 8B/GLM positives.}
\label{tab:repair-cost-recovery}
\end{table}

\paragraph{Repair cost and recovery.} Within the local subsystem, acquisition with calibration fixed raises development Pair by 3.57 [0.89, 7.14] / 6.25 [1.79, 11.61] points (8B/GLM); COCO120 gains are 0.83 [0.00, 2.50] per reader. Brackets give paired 95\% source-bootstrap intervals. In one proposal, calibration adds 37 checks on development and strict fallback adds 25/66 on development/COCO120. Retry adds 304 queries on development and 34/10 draws on development/COCO120, totaling 79/52 draws, without new masks. Historical chroma needs no draw; one/two half-steps add 24/38 queries. RISE's full workflows use 810/806 reasoner calls and 146/141 images, including planning, generation and verification.

\paragraph{Local-only timing.} On a separate sixteen-conversation set, warm local-only processing averages 54.0\,s per turn with Klein and 195.1\,s with Qwen, using two A100-80GB GPUs at batch size one. Timing starts after the shared first turn, includes ordinary editing, reasoning, verification and saving the returned image, and excludes model loading.

\subsection{Independent new-source conversations}
\label{app:coco-independent}
COCO120 contains fifteen validation photographs each for bus, truck, chair, bench, cup, bowl, bottle and vase. Annotation-only selection requires a unique non-crowd target, box area 8--55\%, sides $\geq18\%$ of image dimensions, and instance area $\geq55\%$ of its box. Sources are distinct from the benchmark photographs and randomly ordered before editing. Neutral padding preserves aspect ratio at $512^2$.

A fixed rotation assigns eight colors, then wood for bus, truck and bottle or metal otherwise. Klein uses seed 42 and the development sampler. Externals share T1; internals share $A$. All 120 sources remain, including fallbacks and unknowns.

\label{app:context-preservation}

\subsection{Goal lifetimes and revisions}
\label{app:goal-lifecycle}
Sixteen eight-turn conversations interleave material changes and color revisions. Active and Accumulated goals share the planner/memory and differ in retiring superseded values. Policies share T1, then follow their own histories. Checks use original introducing inputs; intervals cluster sources (Table~\ref{tab:goal-lifecycle-full}).

Scope selection improves all four context measures with positive intervals: object-feature gains are $2.63$ [1.20, 4.18] and background-feature gains $0.58$ [0.22, 1.00] points. Carrier-executor native joint counts tie across scope policies; appearance counts vary by reader.
\subsection{Independent three-turn conversations}
\label{app:independent32}
Each COCO category (bus, chair, cup, truck, car, bicycle, umbrella and clock) contributes four annotation-selected, source-disjoint photographs. Color revisions retain material.

\end{document}